\documentclass[lettersize,journal]{IEEEtran}
\usepackage{amsmath,amsfonts}
\usepackage[ruled, linesnumbered, vlined]{algorithm2e}
\SetNlSty{text}{}{:}
\SetArgSty{upshape}
\usepackage{array}
\newcolumntype{C}[1]{>{\centering\let\newline\\\arraybackslash\hspace{0pt}}m{#1}} 
\usepackage{textcomp}
\usepackage{stfloats}
\usepackage{url}
\usepackage{verbatim}
\usepackage{graphicx}
\usepackage{threeparttable}
\usepackage{colortbl}
\usepackage{array}
\usepackage{hhline}
\usepackage[dvipsnames]{xcolor}
\usepackage{tikz}
\usepackage{cite}
\usepackage{multirow}
\usepackage{pgfplots}
\pgfplotsset{compat=1.8}
\usepgfplotslibrary{statistics}
\usepackage{subcaption}

\usepackage[most]{tcolorbox}
\newtcolorbox{highlighted}{colback=yellow,coltext=red,breakable}

\begin{document}

\title{High-Speed Motion Planning for Aerial Swarms in Unknown and Cluttered Environments}

\author{Charbel Toumieh, Dario Floreano,~\IEEEmembership{Fellow,~IEEE}
\thanks{The authors are with
the Laboratory of Intelligent Systems, Ecole Polytechnique Federale de Lausanne (EPFL), CH1015 Lausanne, Switzerland (e-mail:
charbel.toumieh@epfl.ch, dario.floreano@epfl.ch).}%
\thanks{This work was supported by the Swiss National Science Foundation (SNSF) with grant number 200020\textunderscore212077. 
}%
\thanks{Video: Supplementary Material}%
\thanks{Code: https://github.com/lis-epfl/multi\textunderscore agent\textunderscore pkgs}%
}



\maketitle

\begin{abstract}
Coordinated flight of multiple drones allows to achieve tasks faster such as search and rescue and infrastructure inspection. Thus, pushing the state-of-the-art of aerial swarms in navigation speed and robustness is of tremendous benefit. In particular, being able to account for unexplored/unknown environments when planning trajectories allows for safer flight.
In this work, we propose the first high-speed, decentralized, and synchronous motion planning framework (HDSM) for an aerial swarm that explicitly takes into account the unknown/undiscovered parts of the environment.  
The proposed approach generates an optimized trajectory for each planning agent that avoids obstacles and other planning agents while moving and exploring the environment. The only global information that each agent has is the target location. The generated trajectory is high-speed, safe from unexplored spaces, and brings the agent closer to its goal. 
The proposed method outperforms four recent state-of-the-art methods in success rate (100\% success in reaching the target location), flight speed (97\% faster), and flight time (50\% lower). Finally, the method is validated on a set of Crazyflie nano-drones as a proof of concept.
\end{abstract}


\section{Introduction}
\IEEEPARstart{T}{he} ability of aerial swarms to rapidly fly through cluttered environments while avoiding each other and any other obstacles could result in faster mission accomplishment and increase area coverage in inspection tasks, aerial logistics, or search for rescue missions. However, current approaches to coordinated flight of multiple drones cannot leverage the maximum possible speed of the drones and do not meet the speed conditions for the longest flight distance \cite{bauersfeld2022range}. Although recent work on drone racing achieved human-competitive flight speed through gates \cite{kaufmann2023champion}\cite{song2023reaching}, those trajectory planning algorithms apply to a single drone, assume knowledge of the environment, and do not generalize well to flight in environments with unknown obstacle layouts.
 
\begin{figure}
\centering
\includegraphics[trim={0cm 0cm 0cm 0cm},clip,width=1\linewidth]{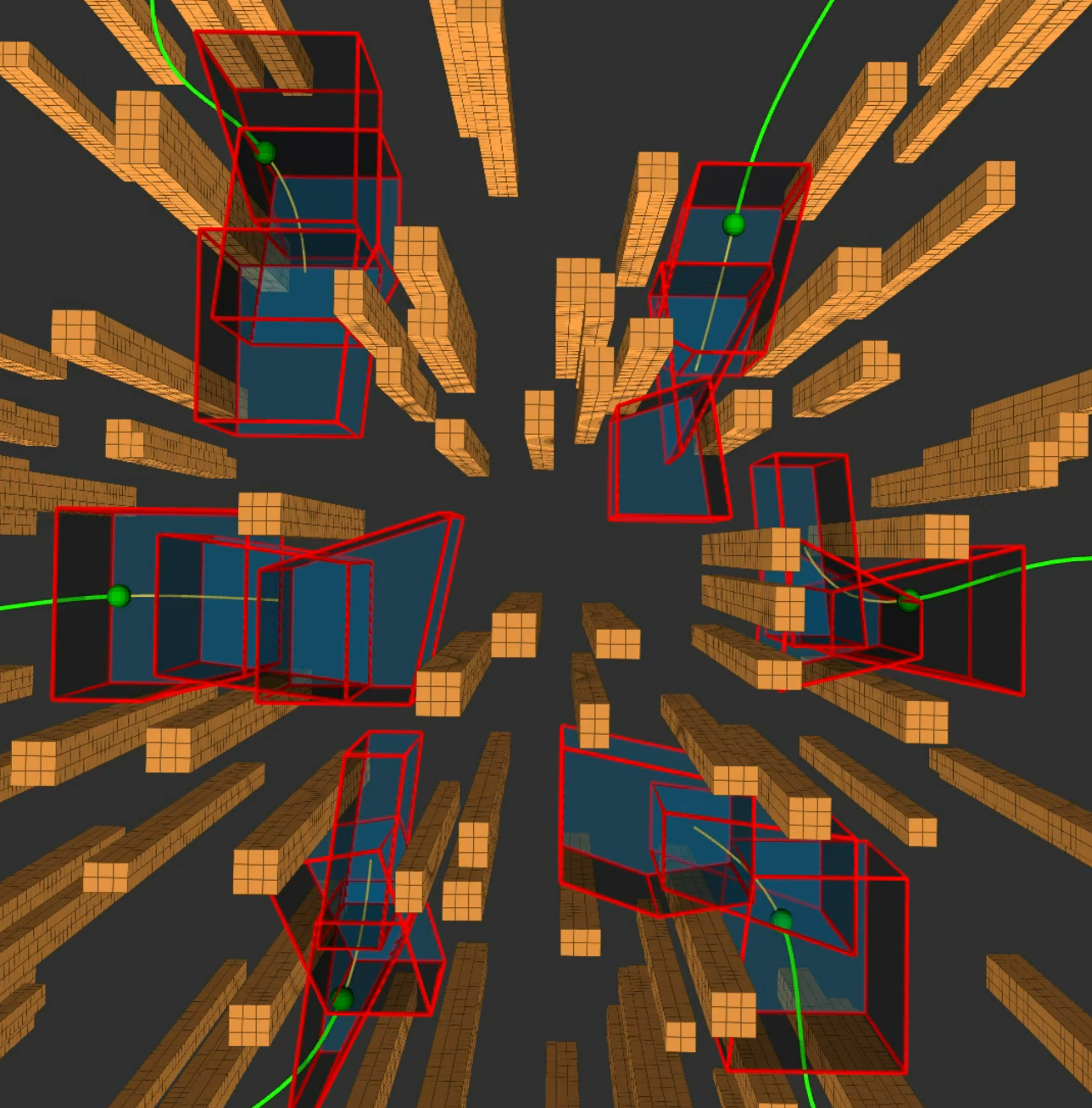}
\caption{Multiple agents (green spheres) moving towards each other in a simulation environment where the obstacles are occupied voxels in orange. The Safe Corridor of each agent is shown in red, the previous positions as a green line, and the predicted future positions (MIQP/MPC solution) as the yellow line.}
\label{fig:sim_example}
\end{figure}

Many works in the literature tackled single-agent trajectory planning in unknown environments \cite{zhou2020ego} \cite{zhou2021raptor} \cite{ren2023online}, but only a few can guarantee collision-free navigation by explicitly taking into account the unknown space. FASTER \cite{tordesillas2020faster} planned 2 trajectories: one that assumes the unknown space to be free (main trajectory), and another that assumes the unknown space to be occupied (backup trajectory). If after sensory measurement updates, it turns out that the unknown space was occupied, the planner switches to the backup trajectory to guarantee safety. The approach is not trivially extendable to multi-agent systems due to its computational cost and inability to readily account for the position of other agents. Another approach \cite{toumieh2020planning} employed Safe Corridors that covered only the free space of the environment and constrained the trajectory inside the safe corridors to guarantee safety. We build on this approach in our work here.

While early methods of trajectory planning of aerial swarm in cluttered environments were centralized \cite{Hoenig2018}, more recent work focused on decentralized approaches where each drone can compute its trajectory based on local information \cite{toumieh2022multi}\cite{tordesillas2020mader}\cite{Zhou2021EGOSwarmAF}, and thus better scale up in computation time and communication range. Consequently, here we discuss the state-of-the-art in decentralized approaches that produce coordinated flight trajectories in cluttered environments (Tab. \ref{tab:sota_overview}). These methods fall into two categories depending on inter-drone communication mode: asynchronous or synchronous.

Asynchronous approaches do not require periodic communication between agents. EGO-Swarm \cite{Zhou2021EGOSwarmAF}, an extension of a gradient-based planner for a single agent \cite{zhou2020ego}, is a recent example of an asynchronous multi-drone planner. However, it cannot handle communication delay between the agents and assumes perfect knowledge of occluded obstacles. EGO-Swarm2 \cite{zhou2021decentralized} uses the MINCO \cite{wang2022geometrically} trajectory parametrization instead of B-Splines \cite{kunoth2018foundations} of EGO-Swarm to produce smoother trajectories and a lower optimization time. However, it cannot handle communication delay and it flies slower in unknown environments \cite{zhou2022swarm}. EDG-Team \cite{hou2022enhanced} improves over EGO-Swarm2 by dealing with deadlocks between multiple agents that pass through narrow gaps. This is achieved by switching the method to a centralized and synchronous planner in dense environments. The method was proven to be more robust to communication delay but could not guarantee collision-free navigation in unknown environments. MADER \cite{tordesillas2020mader} and its delay-robust version RMADER \cite{kondo2022robust} are another family of asynchronous planners. They both use the MINVO basis \cite{tordesillas2022minvo} to generate trajectories that can pass through narrow gaps. Both approaches assume perfect knowledge of the obstacle positions and shapes within a bounding box around each agent and do not account for the unknown part of the environment, which could lead to collisions. PUMA \cite{kondo2023puma} builds on RMADER to make it more robust to unknown environments by pushing the trajectory to be in the field of view of the agent. This, however, does not guarantee safety. DREAM \cite{csenbacslar2023dream} is another asynchronous planning method for aerial swarms that minimizes collision probabilities, but cannot guarantee safety in unknown environments. MRNAV \cite{csenbacslar2023mrnav} was built on top of DREAM to provide collision-free and deadlock-free flight using a centralized, long-horizon planning module. MRNAV retains communication delay robustness of DREAM but still does not guarantee safety in unknown environments. 

\begin{table}
\centering
\caption{Comparison of recent motion planning methods according to key properties: asynchronous or synchronous (\textbf{Async}); decentralized or centralized (\textbf{Decentr.}); handling of communication delay between agents (\textbf{Delay}); explicit handling of unknown parts of the environment (\textbf{Unkn.}).
}
 \begin{threeparttable}
\begin{tabular}{c|c|c|c|c}\hline
 \textbf{Method} & \textbf{Async} & \textbf{Decentr.} & \textbf{Delay} & \textbf{Unkn.} \\ \hhline{=====}
EGO-Swarm \cite{Zhou2021EGOSwarmAF} & \textcolor{Green}{Yes} &  \textcolor{Green}{Yes} & \textcolor{Red}{No} & \textcolor{Red}{No} \\ \hline
EGO-Swarm2 \cite{zhou2021decentralized}\cite{zhou2022swarm} & \textcolor{Green}{Yes} &  \textcolor{Green}{Yes} & \textcolor{Red}{No} & \textcolor{Red}{No} \\ \hline 
EDG-Team \cite{hou2022enhanced} &  \textcolor{Green}{Yes}$\mid$\textcolor{Red}{No}\tnote{1} & \textcolor{Green}{Yes}$\mid$\textcolor{Red}{No}\tnote{1} & \textcolor{Red}{No} & \textcolor{Red}{No} \\ \hline
MADER \cite{tordesillas2020mader} & \textcolor{Green}{Yes} &  \textcolor{Green}{Yes} & \textcolor{Red}{No} & \textcolor{Red}{No} \\ \hline
RMADER \cite{kondo2023robust} & \textcolor{Green}{Yes} &  \textcolor{Green}{Yes} & \textcolor{Green}{Yes} & \textcolor{Red}{No} \\ \hline
PUMA \cite{kondo2023puma} & \textcolor{Green}{Yes} &  \textcolor{Green}{Yes} & \textcolor{Green}{Yes} & \textcolor{Green}{Yes}$\mid$\textcolor{Red}{No}\tnote{2} \\ \hline
DREAM \cite{csenbacslar2023dream} & \textcolor{Green}{Yes} &  \textcolor{Green}{Yes} & \textcolor{Green}{Yes} & \textcolor{Red}{No} \\ \hline
MRNAV \cite{csenbacslar2023mrnav} & \textcolor{Green}{Yes}$\mid$\textcolor{Red}{No}\tnote{3} & \textcolor{Green}{Yes}$\mid$\textcolor{Red}{No}\tnote{3} & \textcolor{Green}{Yes} & \textcolor{Red}{No} \\ \hline
AMSwarmX \cite{adajania2023amswarmx} & \textcolor{Red}{No} &  \textcolor{Green}{Yes} & \textcolor{Red}{No} & \textcolor{Red}{No} \\ \hline
DPDS \cite{soria2021distributed} & \textcolor{Red}{No} &  \textcolor{Green}{Yes} & \textcolor{Red}{No} & \textcolor{Red}{No} \\ \hline
LSC \cite{park2020efficient} & \textcolor{Red}{No} &  \textcolor{Green}{Yes} & \textcolor{Red}{No} & \textcolor{Red}{No} \\ \hline
DLSC \cite{park2023dlsc} & \textcolor{Red}{No} &  \textcolor{Green}{Yes} & \textcolor{Red}{No} & \textcolor{Red}{No} \\ \hline
decMPC \cite{toumieh2022multi} & \textcolor{Red}{No} &  \textcolor{Green}{Yes} & \textcolor{Green}{Yes} & \textcolor{Red}{No} \\ \hline
decMPC2 \cite{toumieh2023decentralized} & \textcolor{Red}{No} &  \textcolor{Green}{Yes} & \textcolor{Green}{Yes} & \textcolor{Red}{No} \\ \hhline{=====}
\textbf{HDSM (proposed)} & \textbf{\textcolor{Red}{No}} &  \textbf{\textcolor{Green}{Yes}} & \textbf{\textcolor{Green}{Yes}} & \textbf{\textcolor{Green}{Yes}} \\ \hline
\end{tabular}
 \begin{tablenotes}
      \item[1] In dense environments, EDG-Team switches to a centralized and synchronous planner that executes joint optimization.
      \item[2] PUMA tries to keep the generated trajectory in the field of view of the camera but cannot guarantee safety.
      \item[3] The long horizon module runs on a centralized system that communicates with the planning robots periodically.
    \end{tablenotes}
\end{threeparttable}
\label{tab:sota_overview}
\end{table}

In contrast to asynchronous methods, synchronous approaches require periodic communication of the trajectories between all agents. While this adds a communication constraint on the swarm, synchronous communication can be utilized to help deal with communication delays passively \cite{toumieh2022multi}. It may face more communication challenges when the number of drones is increased, but those challenges can be attenuated through the use of methods that are robust to packet loss and communication delays such as the method proposed in this work. 

AMSwarmX \cite{adajania2023amswarmx} is a synchronous method that uses Bernstein polynomials as trajectory parametrization. However, it assumes no communication delays between the agents as well as prior knowledge of the environment. The same assumptions are made by DPDS \cite{soria2021distributed}, which uses a discretized trajectory and MPC optimization for obstacle avoidance where obstacles are represented by mathematical functions (e.g. cylinders, paraboloids). Some synchronous planning methods are based on linear safe corridors, which use separating hyperplanes to guarantee collision avoidance with static obstacles and the
other planning agents, such as LSC \cite{Park2022DecentralizedDT} and its extension to dynamic obstacles DLSC \cite{park2023dlsc}. However, these methods do not account for communication delay and assume prior knowledge of the environment.

Here we propose a high-speed decentralized and synchronous motion (HDSM) planning method for aerial swarms that can operate in cluttered and unknown environments with a guarantee of collision-free navigation (Fig. \ref{fig:sim_example}). The method builds on our previous work for aerial swarm motion planning \cite{toumieh2022multi} and its extension to arbitrary communication delays \cite{toumieh2023decentralized}. However, those methods were unable to deal with unknown environments, unlike the proposed approach. Furthermore, the method described here introduces novelties that reduce trajectory lengths, allow drones to fly through narrow gaps, and adapt flight speed to environment density. Taken together, all these improvements substantially increase flight speed while producing collision-free trajectories. When compared to four recent and open-source methods that outperform all other state-of-the-art approaches (Sect. \ref{sect:comparison_sota}), the method described here results in 97\% higher flight speed and 50\% lower flight time with 100\% mission success rate. Finally, while all state-of-the-art approaches (Tab. \ref{tab:sota_overview}) are implemented in ROS1 or Matlab, the method described here takes advantage of ROS2 peer-to-peer communication and other real-time and security features \cite{ros2}.

\begin{figure*}
\centering
\input{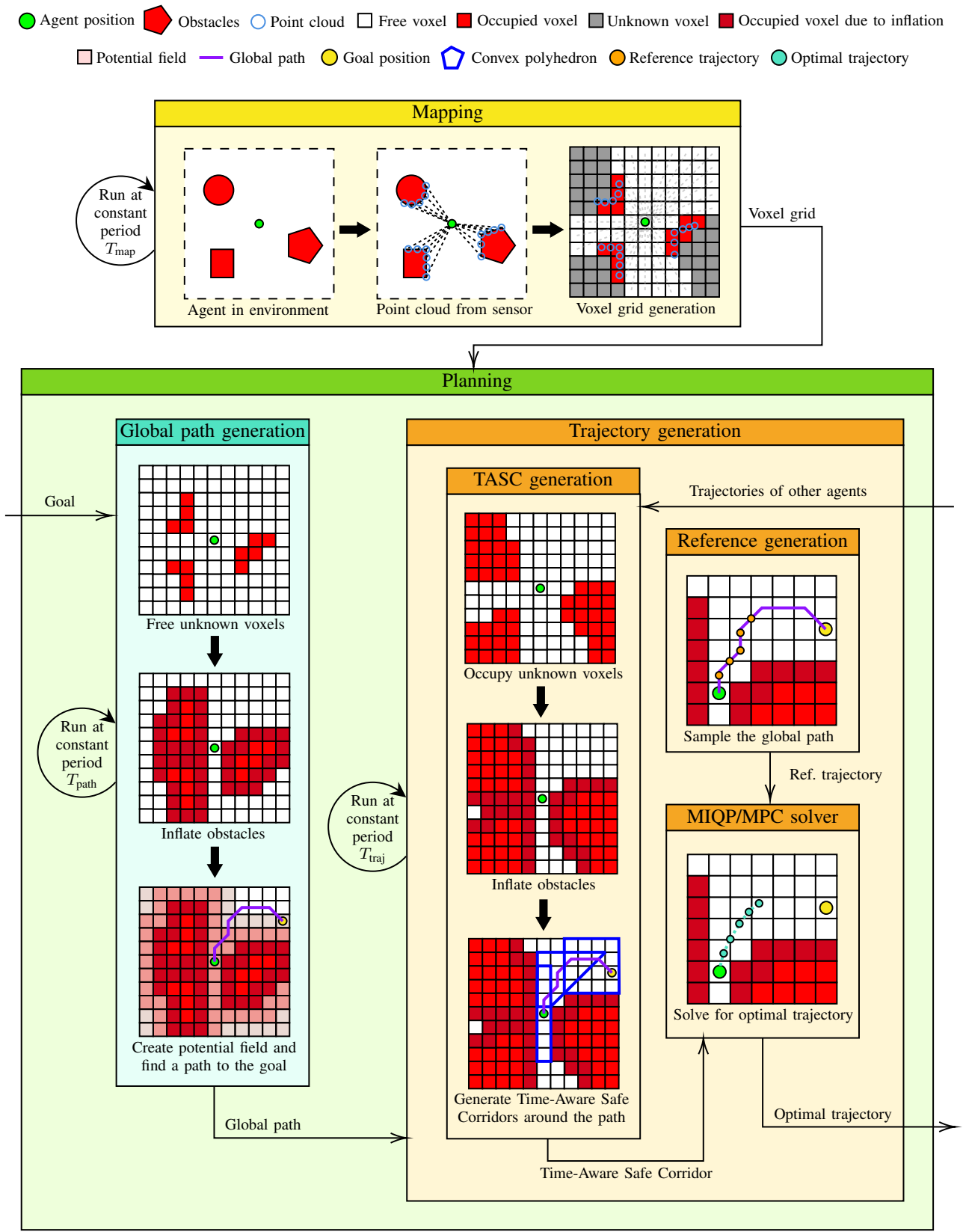}
\caption{The mapping and planning modules are shown in 2D for the sake of clarity. The mapping framework takes a point cloud generated at any time $t$ by sensors and produces a voxel grid with free, occupied, and unknown voxels. It is run at constant period $T_{\text{map}}$. The voxel grid is then fed to the planning framework. The planning framework consists of 2 modules each running at a different frequency and on a different CPU thread: the global path generation module (running at period $T_{\text{path}}$) which takes the voxel grid and generates a global path to the goal; the trajectory generation module (running at period $T_{\text{traj}}$) which takes the voxel grid and the global path to generate a collision-free and dynamically-feasible optimal trajectory.} 
\label{fig:general_pipeline}
\end{figure*}

\section{The HDSM Method}
\subsection{Overview}
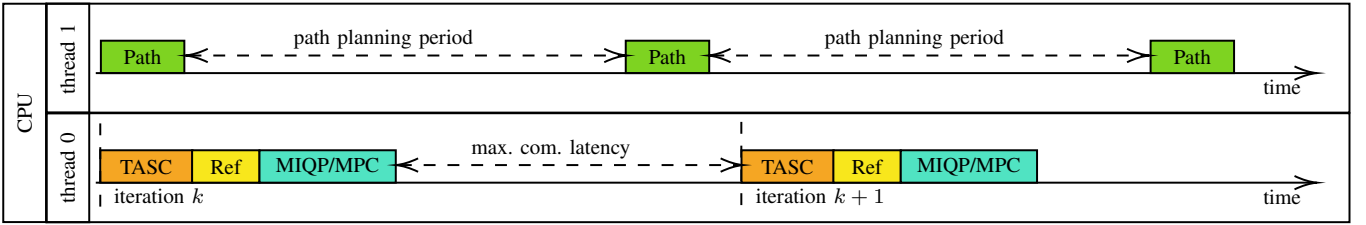
\begin{figure*}
\centering
\tikzset{every picture/.style={line width=0.75pt}} 

\begin{tikzpicture}[x=0.75pt,y=0.75pt,yscale=-1.03,xscale=1.03]

\draw    (79.5,293.62) -- (675.4,293.62) ;
\draw [shift={(677.4,293.62)}, rotate = 180] [color={rgb, 255:red, 0; green, 0; blue, 0 }  ][line width=0.75]    (10.93,-3.29) .. controls (6.95,-1.4) and (3.31,-0.3) .. (0,0) .. controls (3.31,0.3) and (6.95,1.4) .. (10.93,3.29)   ;
\draw  [fill={rgb, 255:red, 126; green, 211; blue, 33 }  ,fill opacity=1 ] (82,224.05) -- (122.87,224.05) -- (122.87,239.93) -- (82,239.93) -- cycle ;

\draw  [fill={rgb, 255:red, 245; green, 166; blue, 35 }  ,fill opacity=1 ] (81.68,277.74) -- (126.86,277.74) -- (126.86,293.62) -- (81.68,293.62) -- cycle ;

\draw  [fill={rgb, 255:red, 248; green, 231; blue, 28 }  ,fill opacity=1 ] (126.67,277.74) -- (159.66,277.74) -- (159.66,293.62) -- (126.67,293.62) -- cycle ;

\draw  [fill={rgb, 255:red, 80; green, 227; blue, 194 }  ,fill opacity=1 ] (159.66,277.74) -- (226.4,277.74) -- (226.4,293.62) -- (159.66,293.62) -- cycle ;

\draw  [dash pattern={on 4.5pt off 4.5pt}]  (81.7,264.58) -- (81.7,305.41) ;
\draw  [dash pattern={on 4.5pt off 4.5pt}]  (395.6,263.83) -- (395.6,304.66) ;
\draw  [fill={rgb, 255:red, 126; green, 211; blue, 33 }  ,fill opacity=1 ] (339,224.05) -- (379.87,224.05) -- (379.87,239.93) -- (339,239.93) -- cycle ;

\draw  [fill={rgb, 255:red, 245; green, 166; blue, 35 }  ,fill opacity=1 ] (395.68,277.74) -- (440.86,277.74) -- (440.86,293.62) -- (395.68,293.62) -- cycle ;

\draw  [fill={rgb, 255:red, 248; green, 231; blue, 28 }  ,fill opacity=1 ] (440.67,277.74) -- (473.66,277.74) -- (473.66,293.62) -- (440.67,293.62) -- cycle ;

\draw  [fill={rgb, 255:red, 80; green, 227; blue, 194 }  ,fill opacity=1 ] (473.66,277.74) -- (540.4,277.74) -- (540.4,293.62) -- (473.66,293.62) -- cycle ;

\draw  [dash pattern={on 4.5pt off 4.5pt}]  (228.4,285) -- (394.4,285) ;
\draw [shift={(396.4,285)}, rotate = 180] [color={rgb, 255:red, 0; green, 0; blue, 0 }  ][line width=0.75]    (10.93,-3.29) .. controls (6.95,-1.4) and (3.31,-0.3) .. (0,0) .. controls (3.31,0.3) and (6.95,1.4) .. (10.93,3.29)   ;
\draw [shift={(226.4,285)}, rotate = 0] [color={rgb, 255:red, 0; green, 0; blue, 0 }  ][line width=0.75]    (10.93,-3.29) .. controls (6.95,-1.4) and (3.31,-0.3) .. (0,0) .. controls (3.31,0.3) and (6.95,1.4) .. (10.93,3.29)   ;
\draw    (79.5,239.93) -- (675.4,239.93) ;
\draw [shift={(677.4,239.93)}, rotate = 180] [color={rgb, 255:red, 0; green, 0; blue, 0 }  ][line width=0.75]    (10.93,-3.29) .. controls (6.95,-1.4) and (3.31,-0.3) .. (0,0) .. controls (3.31,0.3) and (6.95,1.4) .. (10.93,3.29)   ;
\draw  [fill={rgb, 255:red, 126; green, 211; blue, 33 }  ,fill opacity=1 ] (596,224.05) -- (636.87,224.05) -- (636.87,239.93) -- (596,239.93) -- cycle ;

\draw  [dash pattern={on 4.5pt off 4.5pt}]  (125.4,231.31) -- (337.4,231.31) ;
\draw [shift={(339.4,231.31)}, rotate = 180] [color={rgb, 255:red, 0; green, 0; blue, 0 }  ][line width=0.75]    (10.93,-3.29) .. controls (6.95,-1.4) and (3.31,-0.3) .. (0,0) .. controls (3.31,0.3) and (6.95,1.4) .. (10.93,3.29)   ;
\draw [shift={(123.4,231.31)}, rotate = 0] [color={rgb, 255:red, 0; green, 0; blue, 0 }  ][line width=0.75]    (10.93,-3.29) .. controls (6.95,-1.4) and (3.31,-0.3) .. (0,0) .. controls (3.31,0.3) and (6.95,1.4) .. (10.93,3.29)   ;
\draw  [dash pattern={on 4.5pt off 4.5pt}]  (381.4,231.31) -- (593.4,231.31) ;
\draw [shift={(595.4,231.31)}, rotate = 180] [color={rgb, 255:red, 0; green, 0; blue, 0 }  ][line width=0.75]    (10.93,-3.29) .. controls (6.95,-1.4) and (3.31,-0.3) .. (0,0) .. controls (3.31,0.3) and (6.95,1.4) .. (10.93,3.29)   ;
\draw [shift={(379.4,231.31)}, rotate = 0] [color={rgb, 255:red, 0; green, 0; blue, 0 }  ][line width=0.75]    (10.93,-3.29) .. controls (6.95,-1.4) and (3.31,-0.3) .. (0,0) .. controls (3.31,0.3) and (6.95,1.4) .. (10.93,3.29)   ;
\draw   (55.39,205.9) -- (693.39,205.9) -- (693.39,259.29) -- (55.39,259.29) -- cycle ;
\draw   (55.39,259.59) -- (693.39,259.59) -- (693.39,312.98) -- (55.39,312.98) -- cycle ;
\draw   (34.2,312.93) -- (34.2,205.93) -- (55.2,205.93) -- (55.2,312.93) -- cycle ;

\draw   (55.3,259.6) -- (55.3,205.9) -- (76.3,205.9) -- (76.3,259.6) -- cycle ;
\draw   (55.3,313) -- (55.3,259.3) -- (76.3,259.3) -- (76.3,313) -- cycle ;

\draw (507.03,285.68) node   [align=left] {{\footnotesize MIQP/MPC}};
\draw (457.17,285.68) node   [align=left] {{\footnotesize Ref}};
\draw (418.27,285.68) node   [align=left] {{\footnotesize TASC}};
\draw (359.43,231.99) node   [align=left] {{\footnotesize Path}};
\draw (193.03,285.68) node   [align=left] {{\footnotesize MIQP/MPC}};
\draw (143.17,285.68) node   [align=left] {{\footnotesize Ref}};
\draw (104.27,285.68) node   [align=left] {{\footnotesize TASC}};
\draw (102.43,231.99) node   [align=left] {{\footnotesize Path}};
\draw (616.43,231.99) node   [align=left] {{\footnotesize Path}};
\draw (44.7,259.43) node  [font=\footnotesize,rotate=-270] [align=left] {CPU};
\draw (87,294.81) node [anchor=north west][inner sep=0.75pt]   [align=left] {{\footnotesize iteration $\displaystyle \mathnormal{k}$}};
\draw (401.25,294.81) node [anchor=north west][inner sep=0.75pt]   [align=left] {{\footnotesize iteration $\displaystyle \mathnormal{k} +1$ }};
\draw (649.75,295.33) node [anchor=north west][inner sep=0.75pt]   [align=left] {{\footnotesize time}};
\draw (262,270.98) node [anchor=north west][inner sep=0.75pt]  [font=\footnotesize] [align=left] {max. com. latency};
\draw (649.75,241.64) node [anchor=north west][inner sep=0.75pt]   [align=left] {{\footnotesize time}};
\draw (175,217.29) node [anchor=north west][inner sep=0.75pt]  [font=\footnotesize] [align=left] {path planning period};
\draw (435,217.29) node [anchor=north west][inner sep=0.75pt]  [font=\footnotesize] [align=left] {path planning period};
\draw (65.8,232.75) node  [font=\footnotesize,rotate=-269.98] [align=left] {thread 1};
\draw (65.8,286.15) node  [font=\footnotesize,rotate=-269.98] [align=left] {thread 0};
\end{tikzpicture}
\caption{In the proposed framework, the path planning is done on a separate thread (thread 1), and is run at a potentially different frequency than the trajectory planning thread (thread 0).} 
\label{fig:parallel_computing}
\end{figure*}

The HDSM method consists of two modules that run in parallel on each of the $M$ drones that make the aerial swarm: the mapping module and the planning module (Fig. \ref{fig:general_pipeline}).
The mapping module takes in a depth point cloud from sensors and generates a voxel grid representation of the environment. The voxel grid is partitioned into free, occupied, and unknown voxels. The module is updated at a period $T_{\text{map}}$.
The voxel grid is given to the planning module which consists of two sub-modules that run in parallel (Fig. \ref{fig:parallel_computing}). 
The first sub-module takes the voxel grid as input and generates a global path between the current position and the target position at a period $T_{\text{path}}$. 
The second sub-module takes as inputs the voxel grid, the global path, and the trajectories of the other drones, and generates a collision-free and dynamically feasible trajectory at a period $T_{\text{traj}}$. 
Running the path generation and trajectory generation in parallel allows each agent more time to share its trajectory with other drones before the next planning iteration begins (Fig. \ref{fig:parallel_computing}). The periods $T_{\text{map}}$ and $T_{\text{path}}$ are usually chosen equal to the sensor measurement period, whereas the period $T_{\text{traj}}$ is chosen to be smaller than $T_{\text{path}}$.

\subsection{The Mapping Module} \label{sect:mapping_framework}
\begin{figure*}
\centering
\input{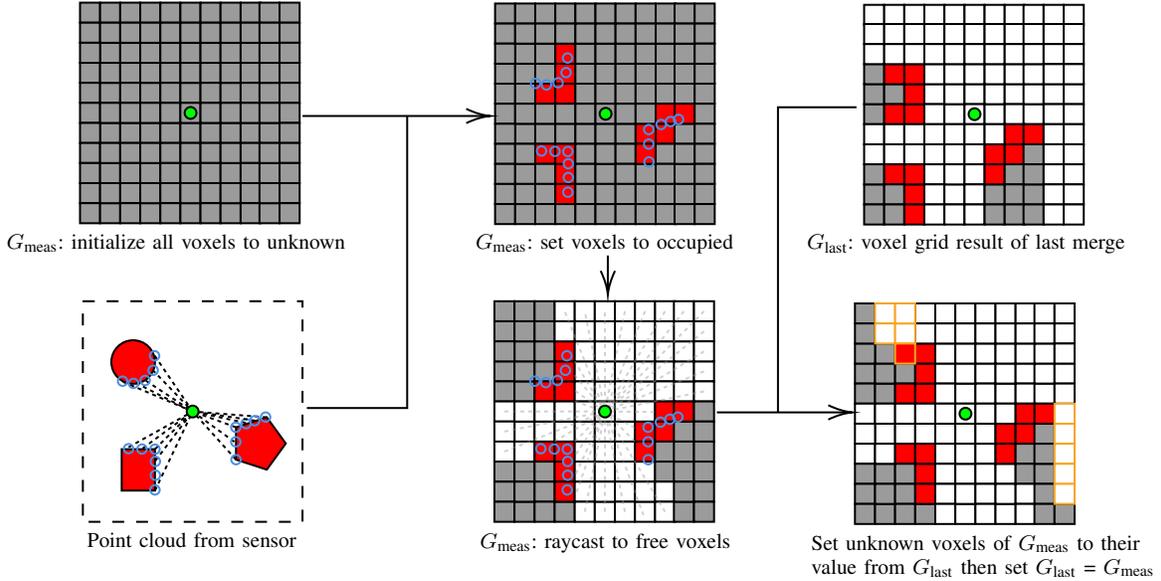}
\caption{The mapping module steps.} 
\label{fig:mapping_framework}
\end{figure*}

The objective of the mapping module is to provide the most up-to-date representation of the local environment in the form of a voxel grid given all the previous sensory measurements. It relies on a depth point cloud from 360-degree sensors, which can be produced by several commercial drones (Skydio \cite{bachrach2021skydio}, DJI \cite{mavic3pro}, e.g.). Fig. \ref{fig:mapping_framework} illustrates the mapping module steps which are taken from \cite{toumieh2020mapping}. It takes as input the most recent point cloud and transforms it into a voxel grid $G_{\text{meas}}$, which can be generated by GPU accelerators for lower latency \cite{toumieh2020mapping}\cite{toumieh2020planning}, which is then merged with the latest computed voxel grid $G_{\text{last}}$. Each grid is a cuboid of fixed dimensions and of fixed voxel size $l_{\text{vox}}$ where each voxel is a cube. 
The user selects the dimensions and voxel size to balance between the time it takes to compute the voxel grid and the extent of the area mapped. The voxel grid's origin coordinates ($x, y, z$) are chosen according to the following rules: each coordinate of the origin is a multiple of $l_{\text{vox}}$ and the voxel at the center of the grid contains the current position of the agent. This makes sure that as the grid moves with the agent, individual voxels can be compared and merged in a one-to-one fashion between $G_{\text{meas}}$ and $G_{\text{last}}$. Every time a new point cloud is produced by the sensors, a grid full of unknown voxels $G_{\text{meas}}$ is generated (measurement grid). 
First, all the voxels that contain at least a point of the point cloud are set to occupied. Raycasting is then performed from the center of the grid to every voxel on the borders of the grid, freeing all the voxels on the raycasted line until an occupied voxel is encountered. Finally, the unknown voxels of $G_{\text{meas}}$ are replaced with their value from $G_{\text{last}}$, and $G_{\text{last}}$ is set to $G_{\text{meas}}$. 
This step can utilize a probabilistic approach (log odds \cite{elfes1989using}) without requiring any modifications to the planning module.

The points in the point cloud that correspond to the other agents can be removed since we know the position of the agents. In this case, we raycast until we hit the drone points, set the voxels containing the drone points to free, and assume everything behind it to be unknown when generating $G_{\text{meas}}$. We can potentially replace the unknown part of the environment (occluded by another drone in the swarm) with information sent from the occluding drone.

\subsection{The global path generation module}
\label{sect:global_planning}
The global path generation module (Fig. \ref{fig:general_pipeline}) takes as input the last voxel grid $G_{\text{last}}$ and the goal position $\boldsymbol{p}_{\text{goal}}$ to generate a path in the free space of the voxel grid. If the goal $\boldsymbol{p}_{\text{goal}}$ is outside the voxel grid, an intermediate goal $\boldsymbol{p}_{\text{goal,inter}}$ is set where the line connecting the agent position and the goal position $\boldsymbol{p}_{\text{goal}}$ intersects the voxel grid border (all border voxels are always left free for positioning of $\boldsymbol{p}_{\text{goal,inter}}$).

\begin{figure*}
\begin{subfigure}{0.2\textwidth}
\centering
\tikzset{every picture/.style={line width=0.75pt}} 

\begin{tikzpicture}[x=0.75pt,y=0.75pt,yscale=-0.7,xscale=0.7]

\draw   (196,110) -- (287,110) -- (287,201) -- (196,201) -- cycle ;
\draw  [fill={rgb, 255:red, 208; green, 2; blue, 27 }  ,fill opacity=1 ] (287,110) -- (378,110) -- (378,201) -- (287,201) -- cycle ;
\draw  [fill={rgb, 255:red, 208; green, 2; blue, 27 }  ,fill opacity=1 ] (196,19) -- (287,19) -- (287,110) -- (196,110) -- cycle ;
\draw   (287,19) -- (378,19) -- (378,110) -- (287,110) -- cycle ;
\draw [line width=3]    (241.5,155.5) -- (328.96,68.04) ;
\draw [shift={(332.5,64.5)}, rotate = 135] [color={rgb, 255:red, 0; green, 0; blue, 0 }  ][line width=3]    (20.77,-6.25) .. controls (13.2,-2.65) and (6.28,-0.57) .. (0,0) .. controls (6.28,0.57) and (13.2,2.66) .. (20.77,6.25)   ;
\draw [color={rgb, 255:red, 126; green, 211; blue, 33 }  ,draw opacity=1 ][line width=3]  [dash pattern={on 7.88pt off 4.5pt}]  (241.5,155.5) -- (332.5,155.5) -- (332.5,69.5) ;
\draw [shift={(332.5,64.5)}, rotate = 90] [color={rgb, 255:red, 126; green, 211; blue, 33 }  ,draw opacity=1 ][line width=3]    (20.77,-6.25) .. controls (13.2,-2.65) and (6.28,-0.57) .. (0,0) .. controls (6.28,0.57) and (13.2,2.66) .. (20.77,6.25)   ;
\draw [color={rgb, 255:red, 126; green, 211; blue, 33 }  ,draw opacity=1 ][line width=3]  [dash pattern={on 7.88pt off 4.5pt}]  (241.5,155.5) -- (241.5,64.5) -- (327.5,64.5) ;
\draw [shift={(332.5,64.5)}, rotate = 180] [color={rgb, 255:red, 126; green, 211; blue, 33 }  ,draw opacity=1 ][line width=3]    (20.77,-6.25) .. controls (13.2,-2.65) and (6.28,-0.57) .. (0,0) .. controls (6.28,0.57) and (13.2,2.66) .. (20.77,6.25)   ;

\end{tikzpicture}
\caption{Traversability check.}
\label{fig:traversability}
\end{subfigure}
\begin{subfigure}{0.8\textwidth}
\centering
\input{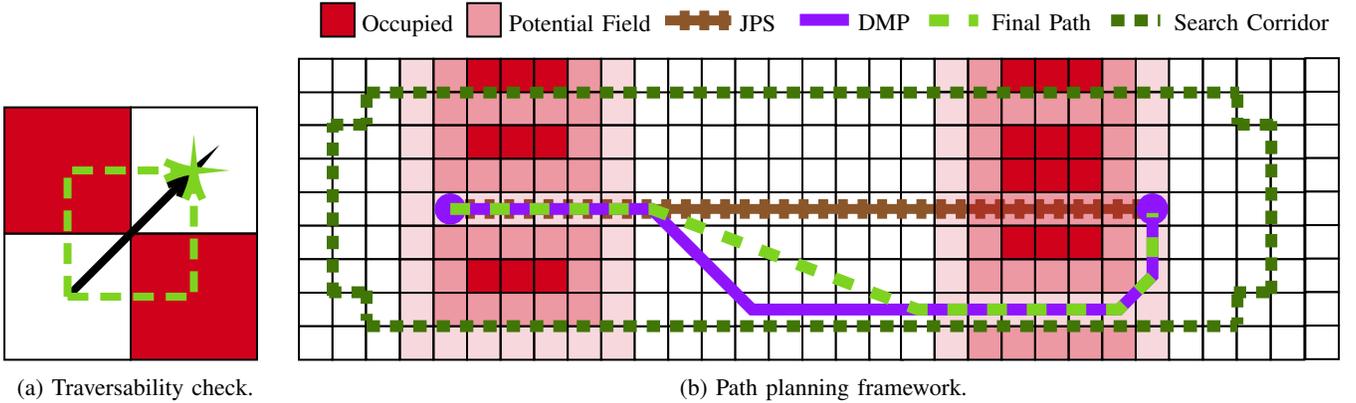}
\caption{Path planning framework.}
\label{fig:path_planning}
\end{subfigure}
\caption{A 2D example of a traversability check (Fig. \ref{fig:traversability}): Green arrows indicate the set of possible paths between adjacent voxels. A diagonal path, indicated by the black arrow, is only considered feasible when one of the (green) paths between adjacent voxels does not result in a collision. The results of JPS, DMP, and the shortened final path in an example 2D environment (Fig. \ref{fig:path_planning}): the final path improves the optimality of DMP while keeping the same margin distance from obstacles.}
\end{figure*}

The module starts from $G_{\text{last}}$ and frees up all unknown voxels to generate $G_{\text{last,free}}$ because if the unknown voxels are assumed to be occupied, the goal may not be reachable when doing the path search. At this point, the module inflates the dimensions of the voxels by taking into account the agent radius (agent modeled as a sphere with radius $r_{\text{agent}}$) and finds a path from the current agent position $\boldsymbol{p}_{\text{curr}}$ to $\boldsymbol{p}_{\text{goal,inter}}$. 

The objective is to find a short path that is also pushed away from obstacles so that there is a safety margin between the drone and the obstacles (the drone will try to follow that path in the trajectory generation module). For this reason, a potential field is created around obstacles to increase the cost of paths passing too close to these obstacles, discouraging their selection during any subsequent path search (see Appendix \ref{app:potential_field} for more details). Doing an A* search with a potential field (denoted as Distance Map Planner (DMP) \cite{jps3d}) over the whole voxel grid would be computationally expensive. For this reason, the shortest path is first found while ignoring the potential field. Then, that shortest path is used to create a small region around it (search corridor). The DMP is then restricted to that region, reducing the number of voxels the DMP search has to consider and thus reducing the computational time. The path generated by DMP is finally shortened for better optimality (Fig. \ref{fig:path_planning} - see Appendix \ref{app:path_shortening}) to get the final global path that the drone will try to follow in the trajectory generation module.

Jump Point Search (JPS) \cite{harabor2011online} is used to first find the shortest path while ignoring the potential field because it offers path optimality guarantees with reduced computation time compared to A*. The search corridor is the union of all the voxels that are within a distance $d_{\text{search}}$ of the JPS path.

DMP or JPS can generate diagonal paths between 2 occupied voxels (Fig. \ref{fig:traversability}). However, since a real drone would not be able to follow that path because there is no empty space between the obstacles, we do not enable such paths.

\subsection{The trajectory generation module}
This module takes as input the last generated global path, voxel grid, and trajectories of other drones, and generates the trajectory executed by the drone (Fig. \ref{fig:general_pipeline}). It runs synchronously on all agents, i.e. all the agents start generating their own trajectory at the same time in a periodic fashion. This requires the clocks of all agents to be synchronized. It is run at a constant period $T_{\text{traj}}$. The objective of this module is to make the drone follow the last generated global path as fast as possible while guaranteeing no collisions with the static obstacles and other agents. This is achieved through 3 sub-modules that run sequentially (Fig. \ref{fig:parallel_computing}): TASC (Time-Aware Safe Corridor) generation, which generates constraints to avoid collisions with obstacles and other drones; reference generation, which samples the global path to generate a reference trajectory for the Model Predictive Controller (MPC) to follow; and Mixed-Integer Quadratic Program/Model Predictive Control (MIQP/MPC) solver, which generates a dynamically feasible trajectory that is constrained in the TASC and that follows the reference trajectory while optimizing for smoothness.

\subsubsection{TASC generation}
\label{sect:tasc}
In this module, a safe corridor is generated around the global path that covers the free space and avoids static obstacles (blue polyhedra in Fig. \ref{fig:general_pipeline}). Then, hyperplanes are added between the agents (in the middle) to split the space into 2 volumes and constrain each agent to one volume to ensure no collisions occur between the agents. The combination of the safe corridor and the added hyperplanes is called the Time-Aware Safe Corridor (TASC). The time awareness comes from the fact that the hyperplanes are generated from the trajectories of the other agents at the previous iteration, and are used to constrain future positions of agents at the current planning iteration. The generation of the TASC is detailed in Appendix \ref{app:tasc_generation}.

\subsubsection{Reference trajectory generation} 
\label{sect:adaptive_speed}
This module's goal is to create a reference trajectory for the MPC/MIQP that has $N$ discretization steps. It samples the most recent global path to produce a reference trajectory for the MPC to follow. Each planning iteration involves sampling $N$ points starting from a reference point $\boldsymbol{p}_{0,\text{ref}}$, which is the agent's location at the very first iteration of the planning algorithm. The sampling progresses along the global path at a speed $v_{\text{samp}}$ to end at $\boldsymbol{p}_{N,\text{ref}}$ (Fig. \ref{fig:general_pipeline}). To ensure efficient navigation, the module employs an algorithm to dynamically modulate $v_{\text{samp}}$, allowing the agent to move faster in open areas and slow down in cluttered spaces.

As the agent moves and progresses along the path, the reference trajectory is updated to reflect this movement, ensuring that it is always pushing the drone along the path and towards the final goal. This update mechanism involves sampling the trajectory at each iteration and adjusting the starting point of the sampling based on the agent's proximity to the points sampled at the previous planning iteration. The full description of the reference trajectory generation process is in Appendix
\ref{app:reference_trajectory}.

\subsubsection{MIQP/MPC solver}
\label{sect:mpc_miqp}
This module takes the reference trajectory and the TASC and generates a dynamically feasible and collision-free trajectory that brings the agent closer to its goal position. The generated trajectory is the result of an MPC/MIQP optimization that constrains the trajectory inside the TASC for collision-free navigation. The cost function of this optimization makes the generated trajectory follow closely the reference trajectory while ensuring a level of smoothness through a jerk cost. The full optimization formulation is derived in Appendix \ref{app:mpc_solver}.

\subsubsection{Communication between agents}
\label{sect:com}
During each planning iteration, every agent requires the previously generated trajectories of other agents to generate the TASC (Sect. \ref{sect:tasc}) and plan safely. This necessitates broadcasting the current trajectory to all other agents immediately after the last step in the trajectory planning module. The trajectory must reach all other agents within $T_{\text{traj}}$ time from the current iteration's start. If the computation time for the current trajectory is $t_{\text{comp}}$, this leaves $t_{\text{lim}} = T_{\text{traj}} - t_{\text{comp}}$ time for the trajectory to reach all other agents before the next planning iteration begins for all agents. Minimizing computation time is essential to accommodate communication latency, hence why the path planning module is done on a different thread (Fig. \ref{fig:parallel_computing}).

Furthermore, if an agent does not receive the trajectory of another nearby agent due to packet loss or because the delay exceeded $t_{\text{lim}}$, it continues executing its previously generated trajectory, ensuring collision-free navigation for all agents in the swarm (see Appendix \ref{app:packet_loss} for more details).
 
\section{Simulation Results} \label{sect:sim_res}

\begin{table*}[ht]
\centering
\caption{Comparison between EGO-Swarm2 (ES2) \cite{Zhou2021EGOSwarmAF}, MADER \cite{tordesillas2020mader}, RMADER \cite{kondo2023robust} and our method HDSM. The comparison is over 100 simulations of 10 agents in a circle of radius 10 m exchanging positions (Fig. \ref{fig:agents_exchange}). The metrics displayed in the table are whether the method is synchronous or asynchronous (Async?), the percentage of generated trajectories  that resulted in a collision, the average number of agent stops per simulation, the jerk cost ($\int_{}{} ||\boldsymbol{j}(t)||^2\mathrm{d}t$), mean flight time, velocity and distance, and the deadlock rate (agents blocking each other indefinitely).}
\begin{threeparttable}
\begin{tabular}{ c | C{1.0cm}|C{1.2cm} |C{1.1cm}|C{1.4cm}|C{1.6cm}|C{1.5cm}|C{1.2cm}}\hline
 Method & Async? & Collision [\%]  & Jerk cost (m\textsuperscript{2}/s\textsuperscript{5}) & Mean flight time (s) & Mean flight velocity (m/s) & Mean flight distance (m) & Deadlock rate [\%] \\ \hhline{========}
AMSwarmX - 2 m/s \cite{adajania2023amswarmx} & No & \textbf{0} & \textbf{25.4} & 12.63 & 1.65 & 20.84 & 0\\ \hline
AMSwarmX - 2.5 m/s \cite{adajania2023amswarmx} & No & 100 & - & - & - & - & -\\ \hline
ES2 - 4 m/s \ \cite{zhou2021decentralized} & Yes & \textbf{0} & 26 & 8.14 & 2.48 & \textbf{20.21} & \textbf{0}\\ \hline
ES2 - 5 m/s \ \cite{zhou2021decentralized} & Yes & \textbf{0} & 80 & 7.65 & 2.65 & 20.28 & \textbf{0}\\ \hline
ES2 - 6 m/s \ \cite{zhou2021decentralized} & Yes & \textbf{0} & 190 & 7.24 & 2.80 & 20.29 & \textbf{0}\\ \hline
ES2 - 7 m/s \ \cite{zhou2021decentralized} & Yes & \textbf{0} & 252 & 7.99 & 2.54 & 20.31 & \textbf{0}\\ \hline
MADER \cite{tordesillas2020mader} & Yes & 5 & 700 & 8.2 & 2.50 & 20.55 & \textbf{0}\\ \hline
RMADER \cite{kondo2023robust} & Yes & \textbf{0} & 2099 & 10.7 & 1.91 & 20.46 & 1\\ \hline

\textbf{HDSM (proposed)} & No & \textbf{0} & 2003 & \textbf{5.61} & \textbf{3.61} & 20.25 & \textbf{0}\\ \hline
\end{tabular}
\end{threeparttable}
\label{tab:comp_free}
\end{table*}

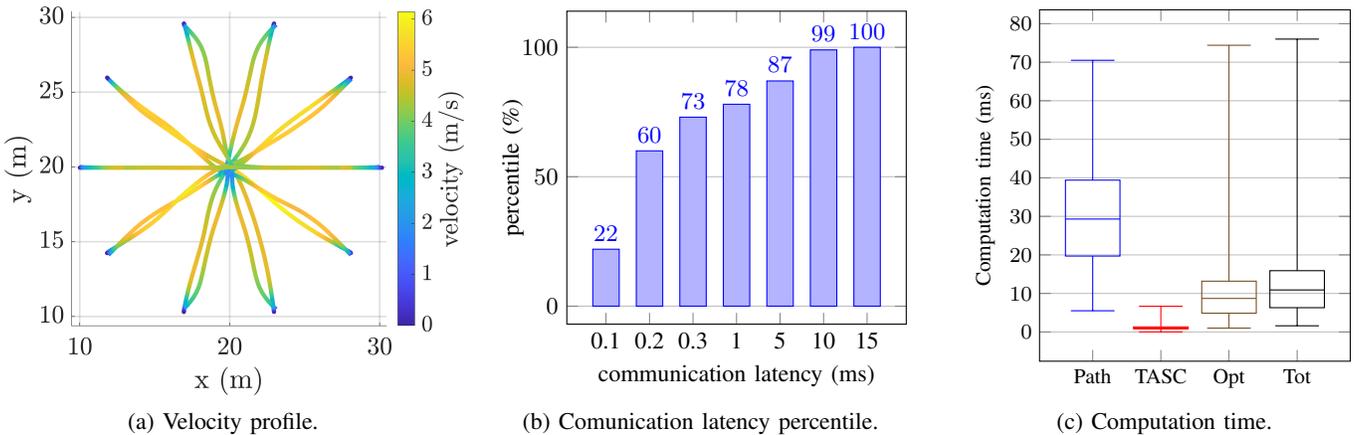
\begin{figure*}
\begin{subfigure}{0.32\textwidth}
\centering
\includegraphics[trim={0cm 0cm 0cm 0cm},clip,width=1.05\linewidth]{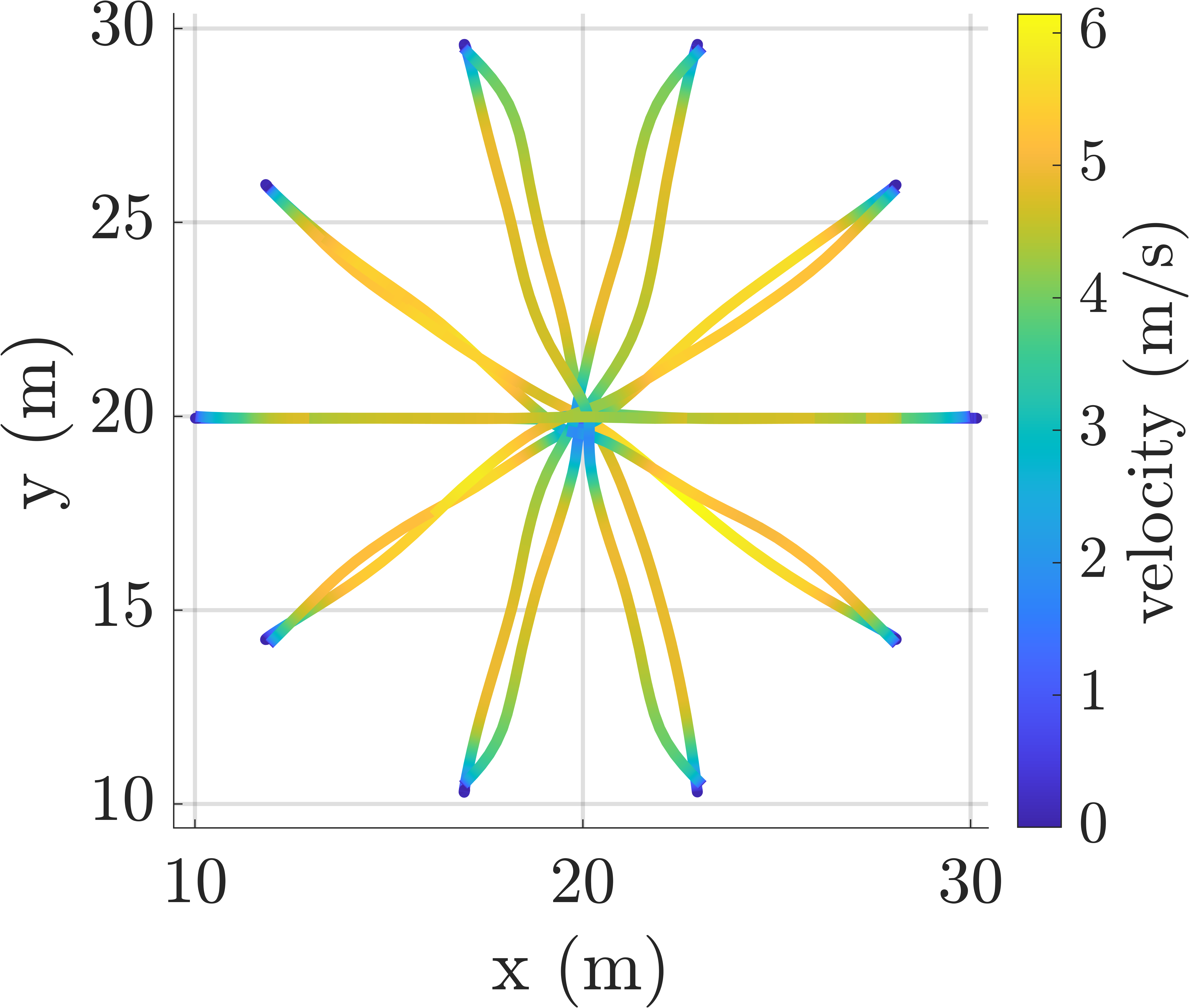}
\caption{Velocity profile.}
\label{fig:agents_exchange}
\end{subfigure}\hfill%
\begin{subfigure}{0.32\textwidth}
\centering
\begin{tikzpicture}
\small
\begin{axis}  
[  width=1.05\textwidth,  
    height=0.99\textwidth,
    ybar,  
    ymin = 7,
    ymax = 100,
    enlargelimits=0.15,  
    ylabel={percentile (\%)}, 
    xlabel={communication latency (ms)},  
    symbolic x coords={0.1, 0.2, 0.3, 1, 5, 10, 15}, 
    xtick=data,  
     nodes near coords, 
    nodes near coords align={vertical},  
    grid=both,
    xmajorgrids=false,
    tick align=inside,
    ylabel style={yshift=-7pt},]
    
\addplot coordinates {(0.1,22) (0.2,60) (0.3,73) (1,78) (5,87) (10,99) (15, 100)};  
\end{axis}  
\end{tikzpicture}
\caption{Comunication latency percentile.}
\label{fig:com_latency}
\end{subfigure}\hfill%
\begin{subfigure}{0.3\textwidth}
\centering
\begin{tikzpicture}
\footnotesize
\begin{axis}[
    width=1.05\textwidth,  
    height=1.15\textwidth, 
    title={},
    ylabel={Computation time (ms)},
    boxplot/draw direction=y,
    boxplot/whisker range=100,
    xtick={1,2,3,4}, 
    ytick={0,10,20,30,40,50,60,70,80},
    xticklabels={Path,TASC,Opt,Tot}, 
    grid=both,
    xmajorgrids=false,
]
\addplot+[ boxplot prepared={
                    lower whisker=5.5,
                    lower quartile=19.7,
                    median=29.3,
                    upper quartile=39.4,
                    upper whisker=70.5
                },
] coordinates {};
\addplot+[ boxplot prepared={
                    lower whisker=0.01,
                    lower quartile=0.8,
                    median=1,
                    upper quartile=1.32,
                    upper whisker=6.68
                },
] coordinates {};
\addplot+[ boxplot prepared={
                    lower whisker=1,
                    lower quartile=4.87,
                    median=8.7,
                    upper quartile=13.18,
                    upper whisker=74.4
                },
] coordinates {};
\addplot+[ boxplot prepared={
                    lower whisker=1.6,
                    lower quartile=6.3,
                    median=10.9,
                    upper quartile=15.9,
                    upper whisker=76
                },
] coordinates {};
\end{axis}
\end{tikzpicture}
\label{fig:all_figs}
\caption{Computation time.}
\label{fig:comp_time}
\end{subfigure}
\caption{The velocity profile (Fig. \ref{fig:agents_exchange}) of 10 agents exchanging their positions using the planner parameters detailed in Sect. \ref{sect:comparison_sota} and Sect \ref{sect:planner_params}. The agents start on a circle of radius 10 m and each agent exchanges its position with the agent that is diametrically opposing it. This simulation is run 100 times and the agents' performance statistics are shown in Tab. \ref{tab:comp_free}. The communication latency percentile (Fig. \ref{fig:com_latency}) between the 10 agents over all simulations (with and without obstacles). All communication latency is below 15 ms. The variation in communication latency is mainly due to the CPU being heavily used. If we only use 5 agents, the communication latency between all the agents would never exceed 10 ms in all the simulations. The computation times (Fig. \ref{fig:comp_time}) of the different steps of our planning algorithm over all simulations (with and without obstacles). Path for the parallel global path generation thread, TASC for the generation of the Time-Aware Safe Corridor, Opt for the optimization time of the MIQP, and Tot for the total computation time of one planning iteration.} 
\label{fig:our_dt}
\end{figure*}

The simulations are run on Intel i9-13900K CPU with a base frequency of 3.0GHz and a turbo boost of 5.8GHz.
ROS2 \cite{ros2} is used for communication between the mapping module and the planning module as well as for communication between the different agents. All methods have access to ground truth states and perfect control. 

The planner is tested in 3 different environments. One free environment and 2 environments with obstacles. The comparison in the free environment is done with AMSwarmX \cite{adajania2023amswarmx}, EGO-Swarm2 (ES2) \cite{zhou2021decentralized}, MADER \cite{tordesillas2020mader} and RMADER \cite{kondo2023robust}. In the environments with obstacles, the comparison is done only with EGO-Swarm2 since it performs considerably better than the other methods.
The comparison is done with 2 versions of our planner. The first version is HDSM where the planner has privileged information about the obstacles i.e. it knows the position and shape of the obstacles within a bounded box around it. Privileged information is required by all the planners used in the comparison. The other version is HDSM* where the planner only sees the obstacles that are in the agent's field of view i.e. the planner is not aware of all the occluded obstacles. This is where our mapping framework comes into play i.e. to determine the unexplored/unseen areas that are unsafe to navigate.
 For comparison between the different methods, the following performance metrics are used: collision [\%] (percentage of simulations that resulted in a collision); flight distance; flight velocity; flight time; acceleration cost ($\int_{}{} ||\boldsymbol{a}(t)||^2\mathrm{d}t$); jerk cost ($\int_{}{} ||\boldsymbol{j}(t)||^2\mathrm{d}t$); success (number of simulation runs where all agents successfully reached their goal position without any collision); deadlock rate [\%] (percentage of generated trajectories that resulted in a deadlock).

\begin{table*}[ht]
\centering
\caption{Comparison between EGO-Swarm2 \cite{zhou2021decentralized} and our planner HDSM/HDSM\textsuperscript{*}. HDSM assumes the planner is aware of all obstacles in a bounding box around the agent. HDSM\textsuperscript{*} is only aware of the obstacles in its field of view and not those that are occluded. HDSM\textsuperscript{*} employs the mapping framework to determine the unknown, free, and occupied parts of the environment. \textbf{HDSM/HDSM* w/o adapt} indicates no speed adaptation according to the density of the environment (use only maximum sampling speed). The comparison is done over 2 environments where the agents have to exchange position in a circular configuration (circle - Fig. \ref{fig:circle_obst}) or traverse a cluttered environment next to each other in a linear fashion (linear - Fig. \ref{fig:linear_env}). The statistics are over 10 randomly generated maps of each environment. The \textbf{mean $\mid$ standard deviation $\mid$ max} of each metric is shown. The better performer between the robust methods (100\% success rate over all simulations) is shown in bold.}
\label{tab:comp_obst}
\begin{threeparttable}
\begin{tabular}{ c | c | c | c | c | c | c | c}\hline
 Exp. & Planner & Succ. & Distance (m) & Velocity (m/s) & Flight time (s) & Acc. cost (m\textsuperscript{2}/s\textsuperscript{3}) & Jerk cost (10\textsuperscript{3}m\textsuperscript{2}/s\textsuperscript{5})\\ \hhline{========}
 \multirow{7}{*}{circle} & ES2 - 4 m/s \cite{zhou2021decentralized} & \textbf{10/10} & 46.1 $\mid$ 0.96 $\mid$ 49.2 & 2.23 $\mid$ 1.24 $\mid$ 4.03 & 20.7 $\mid$ 3.47 $\mid$ 30.1 & \textbf{21.5} $\mid$ 6.84 $\mid$ 49.8 & \textbf{0.16} $\mid$ 0.13 $\mid$ 0.76 \\ 
& ES2 - 5 m/s \cite{zhou2021decentralized} &  \textbf{10/10} & 46.6 $\mid$ 1.11 $\mid$ 52.1 & 2.32 $\mid$ 1.53 $\mid$ 5.04 & 20.1 $\mid$ 4.03 $\mid$ 30.5 & 32.9 $\mid$ 11.9 $\mid$ 79.7 & 1.59 $\mid$ 1.59 $\mid$ 15.6 \\ 
& ES2 - 6 m/s \cite{zhou2021decentralized} &  5/10 & 47.0 $\mid$ 1.67 $\mid$ 53.6 & 2.59 $\mid$ 1.78 $\mid$ 7.84 & 18.2 $\mid$ 4.52 $\mid$ 32.2 & 67.4 $\mid$ 51.5 $\mid$ 240 & 3.65 $\mid$ 3.97 $\mid$ 20.6 \\ 
& \textbf{HDSM} &  \textbf{10/10} & \textbf{45.5} $\mid$ 0.62 $\mid$ 47.5 & \textbf{5.68} $\mid$ 2.45 $\mid$ 10.0 & \textbf{8.01} $\mid$ 0.57 $\mid$ 9.60 & 514 $\mid$ 87.6 $\mid$ 804 & 14.5 $\mid$ 3.36 $\mid$ 23.9 \\ 
& \textbf{HDSM\tnote{*}} &  \textbf{10/10} & 45.9 $\mid$ 0.79 $\mid$ 48.9 & 5.09 $\mid$ 2.27 $\mid$ 9.98 & 9.01 $\mid$ 0.82 $\mid$ 10.6 & 493 $\mid$ 103 $\mid$ 103 & 14.1 $\mid$ 3.40 $\mid$ 21.8 \\ 
\cline{2-8}
& \cellcolor{gray!25}\textbf{HDSM w/o adapt} & \cellcolor{gray!25}10/10 & \cellcolor{gray!25}45.4 $\mid$ 0.60 $\mid$ 47.7 & \cellcolor{gray!25}6.34 $\mid$ 2.76 $\mid$ 10.0 & \cellcolor{gray!25}7.16 $\mid$ 0.33 $\mid$ 7.80 & \cellcolor{gray!25}542 $\mid$ 138 $\mid$ 922 & \cellcolor{gray!25}13.8 $\mid$ 4.53 $\mid$ 26.8 \\ 
& \cellcolor{gray!25}\textbf{HDSM\tnote{*} w/o adapt} & \cellcolor{gray!25}10/10 & \cellcolor{gray!25}45.7 $\mid$ 1.21 $\mid$ 53.1 & \cellcolor{gray!25}6.11 $\mid$ 2.81 $\mid$ 9.99 & \cellcolor{gray!25}7.48 $\mid$ 0.55 $\mid$  9.03 & \cellcolor{gray!25}575 $\mid$ 148 $\mid$ 1098 & \cellcolor{gray!25}15.4 $\mid$ 4.84 $\mid$ 29.7 \\ 
\cline{1-8}
\multirow{6}{*}{linear} & ES2 - 4 m/s \cite{zhou2021decentralized} &  \textbf{10/10} & 101 $\mid$ 3.21 $\mid$ 121 & 2.46 $\mid$ 1.06 $\mid$ 4.01 & 41.1 $\mid$ 4.14 $\mid$ 53.2 & \textbf{35.0} $\mid$ 12.4 $\mid$ 73.0 & \textbf{0.38} $\mid$ 0.50 $\mid$ 4.29 \\ 
& ES2 - 5 m/s \cite{zhou2021decentralized} &  3/10 & 101 $\mid$ 3.07 $\mid$ 109 & 2.68 $\mid$ 1.25 $\mid$ 4.98 & 37.7 $\mid$ 4.49 $\mid$ 45.5 & 41.8 $\mid$ 14.0 $\mid$ 94.1 & 2.80 $\mid$ 2.40 $\mid$ 17.7 \\ 
& \textbf{HDSM} & \textbf{10/10} & \textbf{100} $\mid$ 1.67 $\mid$ 104 & \textbf{6.01} $\mid$ 2.11 $\mid$ 9.78 & \textbf{16.6} $\mid$ 1.04 $\mid$ 18.5 & 851 $\mid$ 154 $\mid$ 1311 & 25.6 $\mid$ 4.59 $\mid$ 38.0 \\ 
& \textbf{HDSM\tnote{*}} & \textbf{10/10} & 102 $\mid$ 3.90 $\mid$ 121 & 5.25 $\mid$ 2.16 $\mid$ 9.71 & 19.4 $\mid$ 1.78 $\mid$ 24.0 & 880 $\mid$ 150 $\mid$ 1320 & 25.8 $\mid$ 4.46 $\mid$ 39.8 \\ 
\cline{2-8}
& \cellcolor{gray!25}\textbf{HDSM w/o adapt} & \cellcolor{gray!25}6/10 & \cellcolor{gray!25}99.6 $\mid$ 1.63 $\mid$ 104 & \cellcolor{gray!25}6.87 $\mid$ 2.30 $\mid$ 9.77 & \cellcolor{gray!25}14.5 $\mid$ 0.85 $\mid$ 16.5 & \cellcolor{gray!25}918 $\mid$ 288 $\mid$ 1508 & \cellcolor{gray!25}25.4 $\mid$ 8.66 $\mid$ 45.5 \\ 
& \cellcolor{gray!25}\textbf{HDSM\tnote{*} w/o adapt} & \cellcolor{gray!25}7/10 & \cellcolor{gray!25}102 $\mid$ 6.10 $\mid$ 131 & \cellcolor{gray!25}6.19 $\mid$ 2.54 $\mid$ 9.69 & \cellcolor{gray!25}16.47 $\mid$ 2.13 $\mid$ 23.1 & \cellcolor{gray!25}1050 $\mid$ 352 $\mid$ 2100 & \cellcolor{gray!25}32.5 $\mid$ 11.2 $\mid$ 67.2 \\ 
\cline{1-8}
\hline
\end{tabular}
\end{threeparttable}
\end{table*}

\subsection{Planner parameters}
\label{sect:planner_params}
The local voxel grid around each agent is of size $20\times 20\times 12$ m and has a voxel size of $0.3$ m.  The following parameters are chosen: $N = 9$, $h = 100$ ms, $v_{\text{samp,min}} = 4.5$ m/s, $v_{\text{samp,max}} = 6$ m/s, $s_d = 0.001$, $s_o = 0.01$, $d_{\text{pot,max}} = 1.5$ m, $d_{\text{search}}  = 1.5$ m, $P_{\text{hor}} = 4$. 
The vertical offset is set equal to the agent radius $z_{\text{offset}} = r_{\text{agent}}$ (the radius will be different for each environment). The period of the mapping module is $T_{\text{map}} = 200$ ms, of the path generation submodule $T_{\text{path}} = 120$ ms, and of the trajectory generation submodule $T_{\text{traj}} = 100$ ms. The starting index for the path planning algorithm is $i_{\text{path,start}} = 6$.

\subsection{Empty environment}
\label{sect:comparison_sota}
 The testing consists of 10 agents in a circular configuration exchanging positions. The circle is of radius 10 m and the agents are positioned in an equidistant way on the circle (Fig. \ref{fig:agents_exchange}). Each agent exchanges its position with the agent that is diametrically opposite. The comparison is done over 100 simulated runs.
 
 The proposed method is compared with AMSwarmX \cite{adajania2023amswarmx}, MADER \cite{tordesillas2020mader}, RMADER \cite{kondo2023robust} and  EGO-Swarm2 (ES2) \cite{zhou2021decentralized}. The maximum acceleration is set to $a_{\text{max}} = 20\ \text{m/s\textsuperscript{2}}$ and the maximum jerk $j_{\text{max}} = 30\ \text{m/s\textsuperscript{3}}$ for all the planners.
For MADER and RMADER, each agent is represented as a bounding box of size $0.25\times0.25\times0.25$ m. For EGO-Swarm2, AMSwarmX, and our planner, each agent is represented as a sphere of diameter $0.25$. 
 The MADER/RMADER values are taken from the results reported by their authors \cite{kondo2023robust} since they have been fine-tuned for this experiment where the maximum speed is set to $v_{\text{max}} = 10$ m/s.

We fine-tuned to the best of our ability the parameters of AMSwarmX, mainly relying on the parameters set by the authors. AMSwarmX results in collisions as soon as the speed limit exceeds 2 m/s. For EGO-Swarm2, we fine-tuned its parameters with a large sensing horizon (10 m) and planning horizon (15 m). Different speed limits are tested: 4, 5, 6, and 7 m/s all shown in Tab. \ref{tab:comp_free}. Note that the average speed performance of the planner improves from 4 m/s until 6 m/s is reached. After that, the average navigation speed decreases while the jerk cost goes up (which is why we stopped at 7 m/s).

Our method outperforms the other methods in mean speed and flight time as shown in Tab. \ref{tab:comp_free}. It is 28.9\% faster than the second-best performer (ES2 - 6 m/s). In terms of flight distance, the difference is negligible between all the methods: the best performer (ES2 - 4 m/s) is only 3\% better than the worst performer (AMSwarmX).

\subsection{Environment with obstacles}
\label{sect:sim_obs}
\begin{figure*}
\begin{subfigure}{0.7\textwidth}
\centering
\includegraphics[trim={0cm 0cm 0cm 0cm},clip,width=1\linewidth]{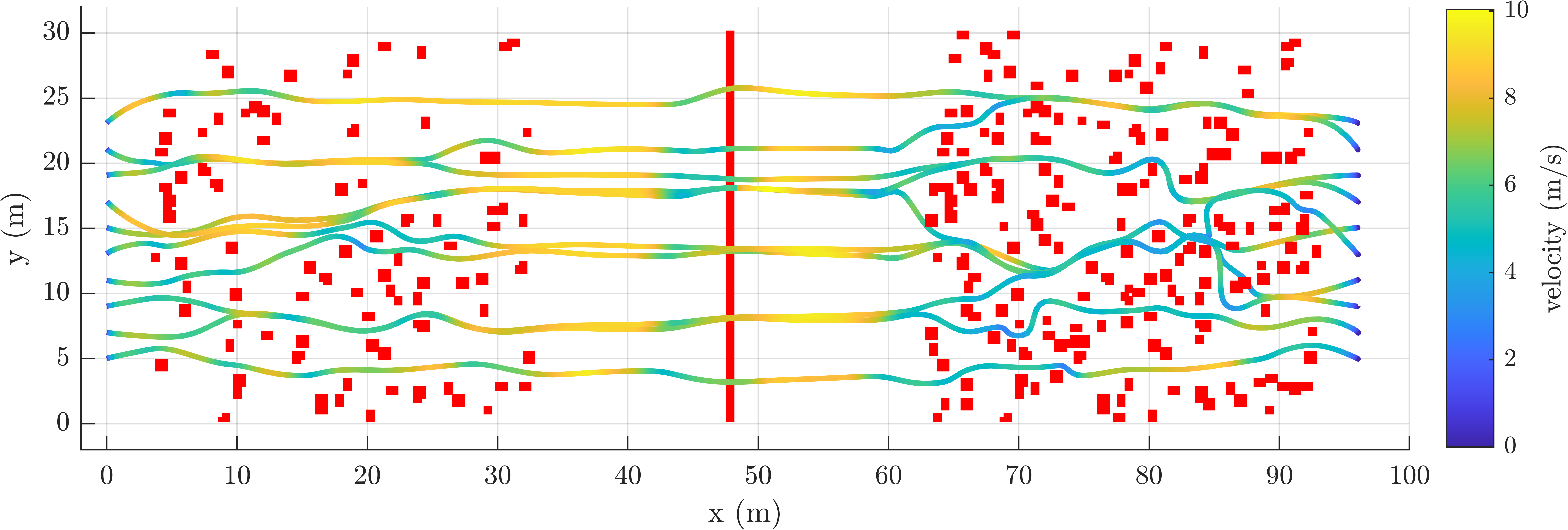}
\caption{Velocity profile.}
\label{fig:linear_env}
\end{subfigure}\hfill
\begin{subfigure}{0.28\textwidth}
\centering
\includegraphics[trim={0cm -3cm 0cm 0cm},clip,width=1\linewidth]{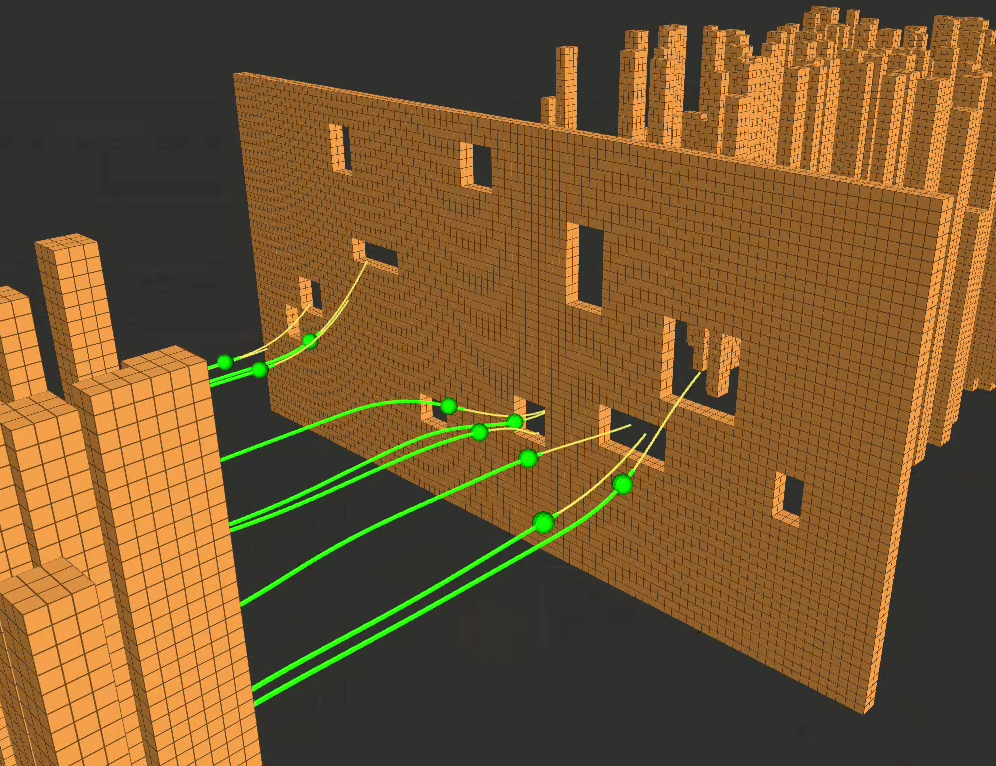}
\caption{Separating wall.}
\label{fig:env_zoom}
\end{subfigure}
\caption{The velocity profile of 10 agents traversing an environment from $x = 0$ m to $x = 96$ m (Fig. \ref{fig:linear_env}). The obstacles are shown in red. The environment consists of 2 areas of different obstacle densities (0.1 obs/m\textsuperscript{2} for $ 3 < x < 33$ and 0.2 obs/m\textsuperscript{2} for $ 63 < x < 93$). These areas are separated by a wall with small openings at varying heights (orange wall in Fig. \ref{fig:env_zoom}). The agents are the green spheres, their previous positions are the green lines and their predicted trajectories (MIQP/MPC solutions) are the yellow lines.}
\label{fig:linear_env_main}
\end{figure*}

\begin{figure}
\centering
\includegraphics[trim={0cm 0cm 0cm 0cm},clip,width=1\linewidth]{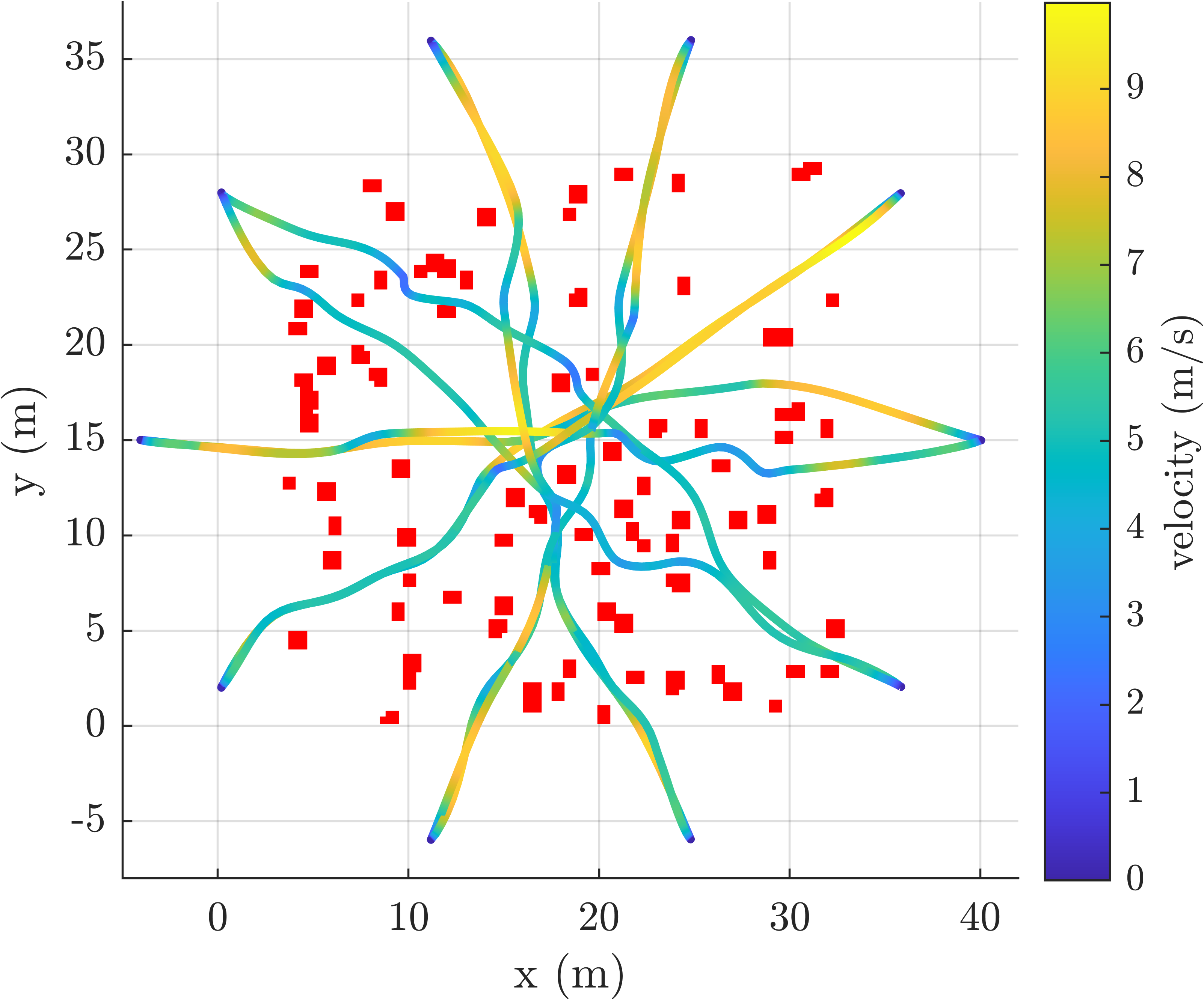}
\caption{The velocity profile of 10 agents in a circular configuration exchanging their positions. The circle is of radius 22 m and contains a forest of cylinders of density of density 0.1 obst/m\textsuperscript{2}.} 
\label{fig:circle_obst}
\end{figure}

Our planner is also tested in 2 environments with obstacles. The first one is a circular exchange: the agents are set up in a circular configuration on a circle of radius 22 m and each agent exchanges its position with the agent that is diametrically opposite. A forest of 90 cylindrical obstacles is generated inside a box of size $30 \times 30 \times 20$ m  that is centered at the circle's center (density 0.1 obst/m\textsuperscript{2}). The positions of the obstacles are generated following a uniform distribution (Fig. \ref{fig:circle_obst}). 

The second environment is linear navigation: the agents start on a line next to each other and then navigate through a cluttered environment to reach their goal. The goal of each agent is its initial position translated 96 m in the positive $x$ direction (Fig. \ref{fig:linear_env}). The cluttered environment consists of 2 cylindrical forests of size $30 \times 30 \times 20$ m separated by a wall with rectangular openings in it (Fig. \ref{fig:env_zoom}). The first forest is centered at $(18, 15, 0)$ and contains 90 cylinders (density 0.1 obst/m\textsuperscript{2}). The second forest is centered at $(78, 15, 0)$ and contains 180 cylinders (density 0.2 obst/m\textsuperscript{2}). The separating wall is 0.6 m thick and contains 11 openings. Its width is 30 m and its height is 15 m.  

The cylinders are all of radius 0.15 m and length 20 m. Their center of gravity is sampled uniformly inside the forest volume. additional environments with obstacles. The radius of each agent is set to 0.3 m (in contrast to 0.125 m in the empty environment) for experimental diversity. We also set $v_{\text{samp,min}} = 5$ m/s and $v_{\text{samp,max}} = 9$ m/s, and compare it with setting a single sampling speed of $9$ m/s (w/o adapt in Tab. \ref{tab:comp_obst}) to showcase the importance of our speed adaptation algorithm \ref{app:reference_trajectory}.

The dynamical limits are set to $a_{\text{max}} = 40$ m/s\textsuperscript{2} and $j_{\text{max}} = 80$ m/s\textsuperscript{3} for EGO-Swarm2 and our planners HDSM/HSDM\textsuperscript{*}. 
The performance of the planner in the considered environments as well as in other environments is shown in the supplementary video. 
The results of 10 simulation runs where the forests of cylinders have been randomized are shown in Tab. \ref{tab:comp_obst}.

HDSM and HDSM\textsuperscript{*} are compared with 3 versions of EGO-Swarm2 where the maximum velocity $v_{\text{max}}$ is set to 4, 5, and 6 m/s. Note that this limit is on the $x$, $y$, and $z$ components of the velocity, and thus the norm of the velocity can get up to $\sqrt{3}\cdot v_{\text{max}}$. In the circle environment, EGO-Swarm2 loses robustness when the maximum speed is set to 6 m/s (5/10 runs where successful): the trajectories start going through the static obstacles. This phenomenon happens when the speed is set to 5 m/s in the linear environment and the collisions happen when agents are trying to traverse the wall. This is because EGO-Swarm2 relies on discretization of the trajectory which can result in a point being on one side of the wall and its subsequent point being on the other side.

In both the circle and linear environments, HDSM and HDSM\textsuperscript{*} largely outperform EGO-Swarm2 in flight velocity and flight time. Note that even though EGO-Swarm2 has privileged information about occluded obstacles in comparison with HDSM\textsuperscript{*}, HDSM\textsuperscript{*} is still 97\% faster in terms of mean flight velocity and has a 50\% lower mean flight time than the best version of EGO-Swarm2 (ES2 - 6 m/s in the circle experiment). In terms of smoothness (acceleration and jerk costs), EGO-Swarm2 outperforms our planner due to its low-speed navigation.  

When speed adaptation is not employed (HDSM/HDSM* w/o adapt in Tab. \ref{tab:comp_free}), the planner exhibits faster performance in low-density environments (circle environment). However, it lacks robustness in cluttered environments (linear environment). The failures observed stem from the optimization solver’s attempt to adhere to an overly ambitious reference trajectory, which makes the generated trajectory very close to obstacles and defeats the purpose of building a potential field. This can lead to local minima problems where the drone becomes stuck, or collisions occur when the solver nudges the drone slightly beyond the Safe Corridor (within the solver’s feasibility tolerance).

\begin{figure*}
\begin{subfigure}{0.5\linewidth}
\centering
\includegraphics[trim={0cm 0.6cm 0cm 1cm},clip,width=1\linewidth]{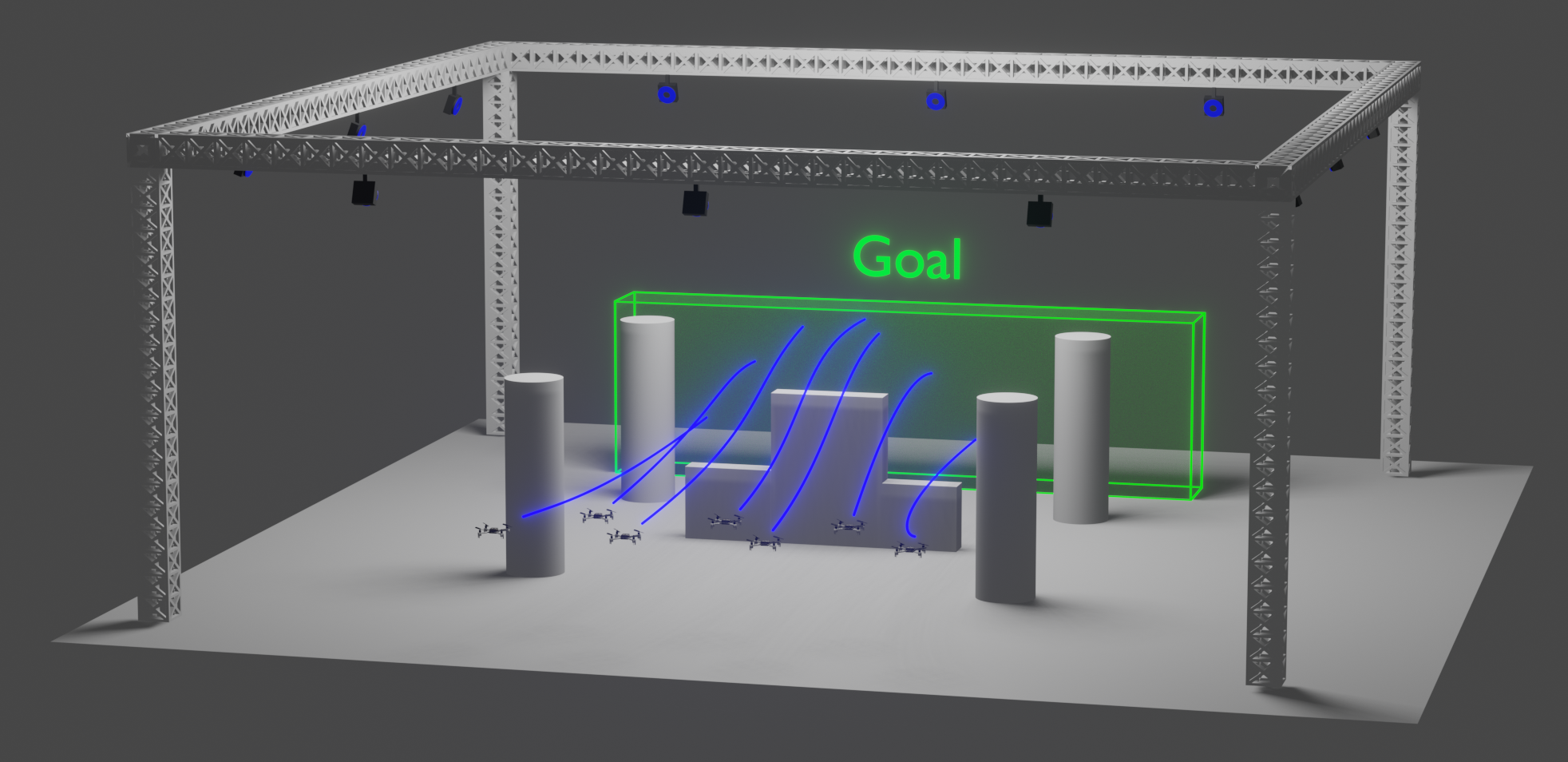}
\caption{Real world experiments schematic.}
\label{fig:realworld_example}
\end{subfigure}
\begin{subfigure}{0.5\linewidth}
\centering
\includegraphics[trim={0cm 0cm 0cm 0cm},clip,width=1\linewidth]{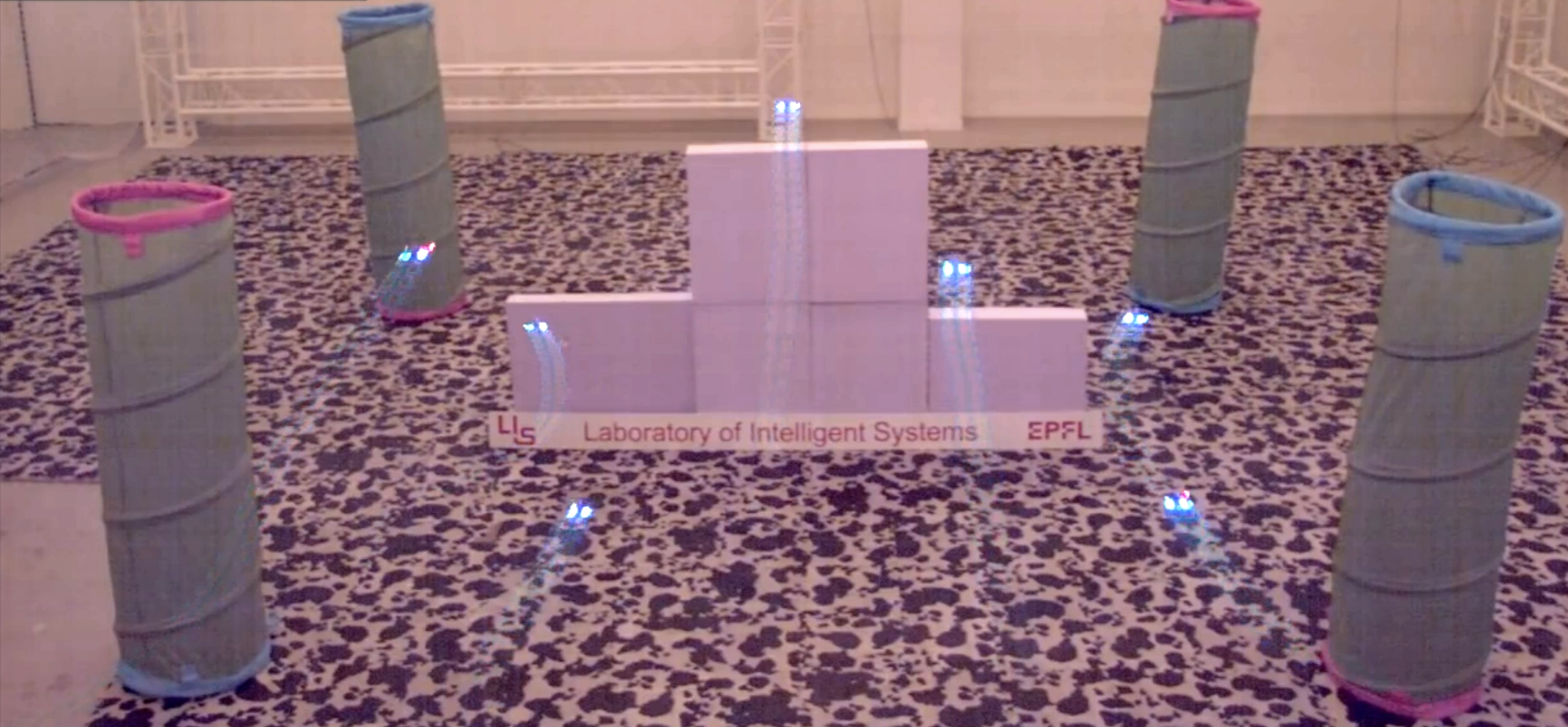}
\caption{Real world experiments picture.}
\label{fig:realworld_cam}
\end{subfigure}
\caption{Hardware experiments: a schematic of the real world experiments with 7 Crazyflies \cite{crazyflie} in the motion capture room (Fig. \ref{fig:realworld_example}). The drones have been increased in size $2\times$ in the figure for better visibility. 7 Crazyflie nano-drones positions during the navigation among obstacles (Fig. \ref{fig:realworld_cam}. The previous positions of each drone are shown with increasing transparency.}
\end{figure*}

\subsection{Computation time and communication latency} \label{sect:comp_time}
The computation time of each step of the planning algorithm over all simulations (with and without obstacles) is shown in Fig. \ref{fig:comp_time} using a boxplot showing the median (line inside the box), 25\textsuperscript{th} and 75\textsuperscript{th} percentile (bottom and upper limits of the box), and the min and max (bottom and upper whiskers). The communication latency percentile is shown in Fig. \ref{fig:com_latency}. The computation time for the local reference generation step is negligible compared to the other steps, and thus not shown separately in Fig. \ref{fig:comp_time}. Since 100\% of all communication delay is less than 15 ms, and our total computation time (without the path planning step which is run in parallel) never exceeds 76 ms, safety is guaranteed because $76 + 15 < T_{\text{traj}} = 100$ ms.
 The mean computation time of our planner is 12 ms and the mean for EGO-Swarm2 is 1.6 ms. The computation time of the mapping framework over all simulations that require raycasting (HDSM\textsuperscript{*}) has a mean of 17.6 ms, a standard deviation of 3.9 ms, and a max of 35.3 ms. 

\section{Hardware experiments}
\label{sect:hardware}

The dynamic feasibility of our planner is tested using 7 nano-drones: the Crazyflies \cite{crazyflie}. The drones start in the formation shown in Fig. \ref{fig:realworld_example}, with a 0.9 m distance between them. They then proceed to traverse the environment to the goal position which is their current position translated 6 m forward. After reaching their goal, the drones traverse the environment again in the opposite direction and go back to their initial positions. Due to the computational limitations of the Crazyflie, the planner is run on a separate PC and sends the commands to the drones to execute using CrazySwarm2 \cite{preiss2017crazyswarm}. The positions of the obstacles in the test environment (Fig. \ref{fig:realworld_example}) are assumed to be known beforehand since the sensing capabilities of the Crazyflie are limited. They are placed in the real world for symbolic purposes (Fig. \ref{fig:realworld_cam}).

The dynamics are limited to the following to adhere to the drone's dynamical limitations and the built-in controller performance: $v_{\text{max}} = 1.5$ m/s, $a_{\text{max}} = 3$ m/s\textsuperscript{2}, $j_{\text{max}} = 4$ m/s\textsuperscript{3}, $v_{\text{samp,min}} = 1.5$ m/s, $v_{\text{samp,max}} = 1.5$ m/s. The other planner parameters are the same as the ones presented in Sect. \ref{sect:planner_params}. The Crazyflie has a radius of $7$ cm radius, but the collision radius is set to $r_{\text{agent}} = 0.3$ m, and the vertical offset to avoid the downwash of other drones is set to $z_{\text{offset}} = 0.6$ m. The large safety radius is to avoid interactions between the airflow/downwash of the drones, which can make the drones unstable.  The Mellinger controller \cite{mellinger2011minimum} is used. However, one could reduce the safety radius as well as the vertical offset by employing recent advances in control for swarms such as Neural-Swarm2 \cite{shi2021neural}.

The voxel size is set to $0.2$ m and the obstacles are inflated by one voxel. The potential distance is set to $d_{\text{pot,max}} = 0.4$ m, and $d_{\text{search}}  = 0.6$ m. The drones were able to execute the trajectories without any crashes (going to the goal area and back to the initial position 2 times). The mean tracking error was 6.8 cm, the standard deviation 2.5 cm, and the maximum tracking error was 10.4 cm. Since the distance between the center of gravity of the drone and the obstacles will always be bigger than 20 cm (because the obstacles are inflated by one voxel), safety is guaranteed. This is because the sum of the radius (7 cm) and the maximum tracking error (10.4 cm) is smaller than 20 cm. This demonstrates to a certain extent that the trajectories generated by our planner are dynamically feasible enough for real-world operations. The performance of the drones is shown in the video.

\section{Limitations}
Dynamic obstacles are not considered in the framework. They can be added by creating a larger potential field around them in the direction of their motion or their reachable space.  Limitations of the proposed framework also include the lack of a mechanism to deal with deadlocks that happen when multiple agents are passing through an extremely narrow gap (especially when the agents are coming from opposing sides of the gap). Another notable limitation is dealing with CPU clock drift, which can lead to unsynchronized trajectories and thus, collisions. Modern CPU clocks have a drift of 100 parts per million (ppm) \cite{li2020sundial}. This means if the clocks are synchronized at the start of a mission, they will deviate by 180 ms after 30 minutes of operation (which is much larger than the planning period of 100 ms). This would require periodic synchronization of the clocks, or using clocks with lower drift (0.1 ppm) which makes a single synchronization at the start of the mission sufficient for safe operations.

In terms of planning strategy, one could choose to adopt a similar strategy to FASTER \cite{tordesillas2020faster}: plan a main trajectory where the unknown space is assumed to be free, and a backup trajectory where the unknown space is assumed to be occupied. In case the space turns out to be occupied after a sensory update, the planner switches to the backup trajectory to guarantee safety. 
 While this approach can result in faster trajectories \cite{tordesillas2020faster}, it also necessitates double the computation time at each iteration since we are generating 2 trajectories instead of 1. It also increases the number of constraints in the optimization stage since each agent must now account for 2 possible trajectories that another agent can take. We choose to generate only one trajectory to minimize computation time and make our method deployable on more computationally constrained systems. Adopting the main/backup trajectory approach can be added in agents with powerful compute.

\section{Conclusion}
In this work, a new framework for high-speed aerial swarm planning in unknown and cluttered environments is presented. To the best of our knowledge, it is the only framework that explicitly takes into account the unknown space of the environment and can guarantee safety in unknown environments. The method is compared to 4 state-of-the-art methods in simulation and is shown to outperform them in multiple metrics such as speed (97\% faster) and flight time (50\% lower). Finally, it is tested in the real world on hardware to show the feasibility of the generated trajectories.

\section*{Appendix}
\subsection{Path planning}
\subsubsection{Potential field generation}
\label{app:potential_field}
The potential field is created around the obstacles to push the path generated by DMP away from them. In the voxel grid, occupied voxels are assigned a value of 100, free voxels are assigned a 0 value, and voxels within a given distance $d_{\text{pot,max}}$ from an obstacle are assigned an intermediate value $o_{\text{voxel}}$:
\begin{gather}
o_{\text{voxel}} = 100 \cdot (1 - \dfrac{d_{\text{voxel}}}{d_{\text{pot,max}}})^4 \label{eqn:potential}
\end{gather}
where $d_{\text{voxel}}$ represents the distance from the voxel center to the nearest occupied voxel. The generation of the potential field is done by going over every occupied voxel and setting the voxels within $d_{\text{pot,max}}$ of the occupied voxel to the maximum between their current value and the new value associated with their distance to the occupied voxel $o_{\text{voxel}}$. By taking the maximum we guarantee that the value of each voxel of the potential field will always be the one computed from the closest occupied voxel after going over all the occupied voxels. 

\subsubsection{Path shortening}
\label{app:path_shortening}
The DMP path can be optimized in terms of length by shortening it. The approach we take to shorten it is the following (Alg. \ref{alg:path_shortening}): we walk through the path until we find a point (index $i_{\text{start}}$) that is free and not in a potential field (lines 1-3). Once we find a free voxel/point of the path, we find the furthest subsequent point (index $i_{\text{end}}$) that is also free and whose segment with the initial point does not intersect any obstacles or potential field voxels (lines 4-13). We then remove the points between  $i_{\text{start}}$ and $i_{\text{end}}$ from the path to shorten it (line 14). We continue walking along the path starting from the next point (line 15) in this fashion until we reach the endpoint of the path. An example of the result is shown in dashed green in Fig. \ref{fig:path_planning}.

\begin{algorithm}
\DontPrintSemicolon
\caption{Global path shortening}
\label{alg:path_shortening}
$i = 0$\;
\While{$i + 1 < $ size(path)}{
\If{IsFree(path[$i$])}{ 
 $i_{\text{start}} = i$\; 
 $i_{\text{end}} = i$\; 
 $j = i + 1$\;
 \While{$j < \text{size(path)}$}{
 line\_clear = IsLineClear(path[$i$], path[$j$])\;
\uIf {$\text{IsFree(path[$j$]}$) \textbf{and} line\_clear}{ 
 $i_{\text{end}} = j$\;
 $j = j+1$\;
 }
\Else{ 
 \textbf{break}\;
}
}
 RemovePointsFromPath(path, $i_{\text{start}}$, $i_{\text{end}}$)\;
}
 $i = i + 1$\;
}
\end{algorithm}

\subsection{Trajectory generation}
\subsubsection{TASC generation}
\label{app:tasc_generation}

\begin{figure*}
\begin{subfigure}{0.25\textwidth}
\centering
\includegraphics[trim={0cm 0cm 0cm 0cm},clip,width=0.9\linewidth]{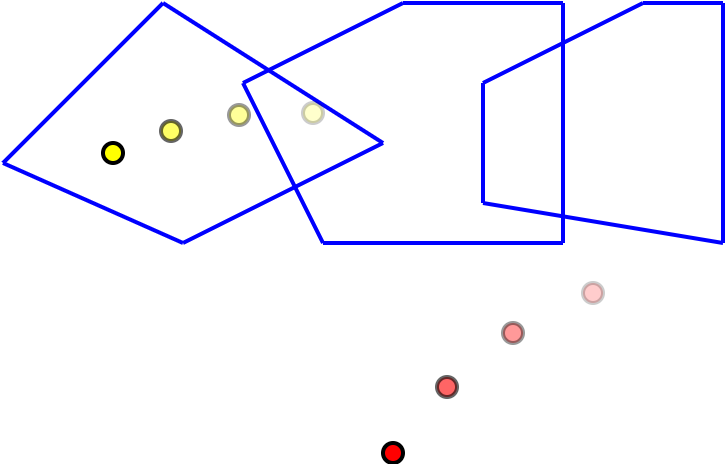}
\caption{$SC_{\text{raw}}$ avoids static obstacles}
\label{fig:tsc_0}
\end{subfigure}
\begin{subfigure}{0.245\textwidth}
\centering
\includegraphics[trim={0cm 0cm 0cm 0cm},clip,width=0.9\linewidth]{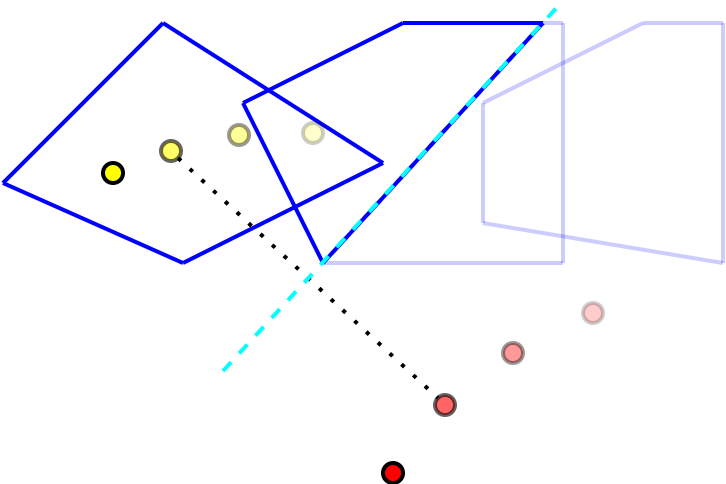}
\caption{Generation of $SC_0$}
\label{fig:tsc_1}
\end{subfigure}
\begin{subfigure}{0.245\textwidth}
\centering
\includegraphics[trim={0cm 0cm 0cm 0cm},clip,width=0.9\linewidth]{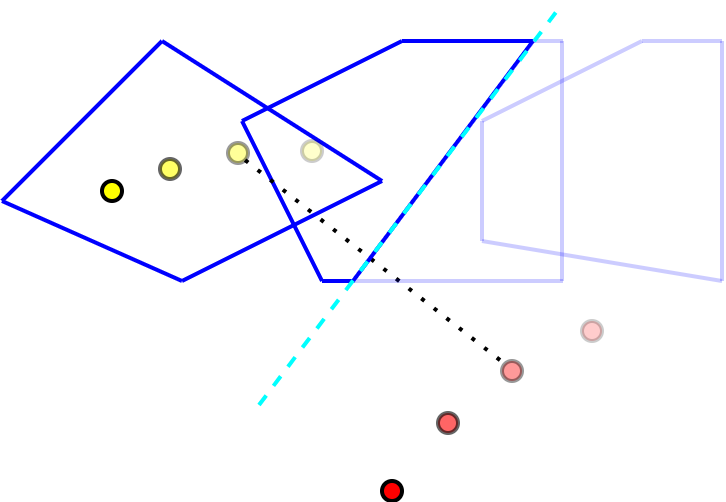}
\caption{Generation of $SC_1$}
\label{fig:tsc_2}
\end{subfigure}
\begin{subfigure}{0.245\textwidth}
\centering
\includegraphics[trim={0cm 0cm 0cm 0cm},clip,width=0.9\linewidth]{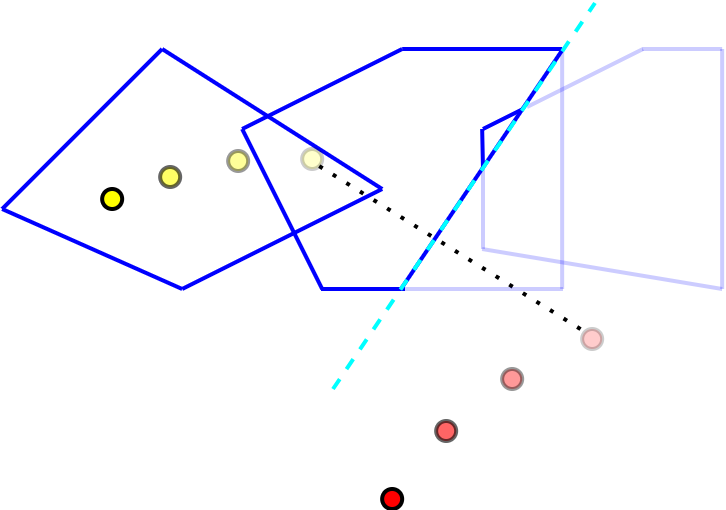}
\caption{Generation of $SC_2$}
\label{fig:tsc_3}
\end{subfigure}

\caption{The generation of a Time-Aware Safe Corridor from $SC_{\text{raw}}$ which only avoids static obstacles \cite{toumieh2022multi}. The predicted positions of the planning agent by the MPC are shown as yellow circles. The predicted positions of the other agent that the planning agent has to avoid are shown as red circles. The positions of each agent become more transparent as we move forward in time. We start from the second position of the trajectory and generate a separating hyperplane that is added as a constraint to each polyhedron of $SC_{\text{raw}}$. This would result in $SC_0$. We do the same procedure for the second and third positions of the trajectory to generate $SC_1$ and $SC_2$. The collection of $SC_0$, $SC_1$, and $SC_2$ is what we denote by Time-Aware Safe Corridor (TASC).}
\label{fig:tsc}
\end{figure*}

The TASC generation consists of first generating a safe corridor that avoids static obstacles and then adding hyperplanes to ensure inta-agent collision avoidance. To generate the safe corridor, we first take the last generated voxel grid by the planning framework $G_{\text{last}}$ and set all the unknown voxels to occupied to generate $G_{\text{last,occ}}$. We then generate a safe corridor that covers only the free space as follows: we first start from the current position of the agent and generate a polyhedron around it. Then, we walk along the global path with a small step ($l_{\text{vox}}/10)$ until we are outside the polyhedron. Once we arrive at the point outside the polyhedron, we inflate a polyhedron around that point. We then continue walking on the global path until we reach a point outside the current polyhedron and we inflate a polyhedron around it. We continue with this procedure until we reach the maximum number of polyhedra $P_{\text{hor}}$ (polyhedron horizon) which is fixed by the user. The generated safe corridor is denoted by $SC_{\text{raw}}$.

Many methods exist to inflate a convex polyhedron around a seed point such as \cite{liu2017planning}\cite{toumieh2020convex}\cite{toumieh2022shape}. When the seed point/voxel around which we want to inflate a polyhedron is restricted in a given direction, we employ \cite{toumieh2022shape} for generating the polyhedron due to its superior performance in cluttered environments (more covered volume). We consider the voxel to be restricted and use \cite{toumieh2022shape} when either the top and bottom voxels are both occupied or the forward and backward are occupied or the left and right are occupied. Otherwise, we employ \cite{toumieh2020convex} to generate the polyhedron. 

We then add a hyperplane between every 2 agents where each agent is constrained to one side of the hyperplane, thus ensuring intra-agent collision avoidance.
The hyperplanes are generated from the trajectory generated at the last planning iteration $\boldsymbol{T}_{\text{last}}$ of the agent as well as the trajectories of the other agents generated at the last planning iteration. We denote by $\boldsymbol{p}_{i}$ the predicted $i^\text{th}$ position of the agent by the optimal trajectory. This means that the agent should reach this point after $i\cdot h$ time, where $h$ is the discretization step of the dynamics in the MPC (Sect. \ref{sect:mpc_miqp}). The hyperplane generation is done between the planning agent and each other agent as follows: for every discrete point $\boldsymbol{p}_{i,\text{plan}}$ in the planning agent trajectory, we take the corresponding point $\boldsymbol{p}_{i,\text{other}}$ of the other agent we are trying to avoid and generate a hyperplane between those points. We define $\boldsymbol{v}_{\text{norm}}$ as the unit vector pointing from $\boldsymbol{p}_{i,\text{other}}$ to $\boldsymbol{p}_{i,\text{plan}}$. The generation of the hyperplane is done as follows: we take the middle point between $\boldsymbol{p}_{i,\text{plan}}$ and $\boldsymbol{p}_{i,\text{other}}$ and then move towards $\boldsymbol{p}_{i,\text{plan}}$ in the direction of $\boldsymbol{v}_{\text{norm}}$ by a distance $d_{\text{offset}}$ which is computed according to an ellipsoid collision model that is used to avoid the down-wash effects of the propellers (the sphere model is elongated in the $z$ direction by $z_{\text{offset}}$). After we move by the distance $d_{\text{offset}}$, we arrive at the point $\boldsymbol{p}_{\text{hyp}}$. The separating hyperplane is thus defined by the point through which it passes $\boldsymbol{p}_{\text{hyp}}$ and its normal vector $\boldsymbol{v}_{\text{norm}}$. The planning agent is then constrained to the side of the hyperplane that it is already in. 

This procedure is done between $\boldsymbol{p}_{i,\text{plan}}$ and every position $\boldsymbol{p}_{i,\text{other}}$ of the other $M-1$ agents. This would generate $M-1$ hyperplane constraints for the planning agent. These constraints are added to each polyhedron of the safe corridor $SC_{\text{raw}}$ (Fig. \ref{fig:tsc}) to generate a new time-local safe corridor $SC_k$ (where $k = i-1$). The generation of $SC_k$ is done for each position $i$ of the last generated trajectory of the planning agent $\boldsymbol{T}_{\text{last}}$ except for the first position $i = 0$. This would result in $N-1$ time-local safe corridors $SC_k$, the collection of which we call Time-Aware Safe Corridor (TASC). The time awareness comes from the fact that each future position predicted by the MPC at the current planning iteration will be constrained to 2 time-local safe corridors. For example, the segment formed by $\boldsymbol{p}_0$ and $\boldsymbol{p}_1$ will be constrained inside $SC_0$, the one formed by $\boldsymbol{p}_1$ and $\boldsymbol{p}_2$ will be constrained inside $SC_1$, and so on. And so, $\boldsymbol{p}_1$ will be constrained inside both $SC_0$ and $SC_1$.

\subsubsection{Reference trajectory generation}
\label{app:reference_trajectory}
The objective of this module is to generate a reference trajectory for the MPC/MIQP which has $N$ discretization steps and a time step $h$. The module takes as input the last generated path from the global path planning module, samples it, and outputs the reference trajectory that the MPC will try to follow. At each iteration $l$, $N$ points are sampled using a starting point  $\boldsymbol{p}^{l}_{0,\text{ref}}$. At the start of a navigation mission, this starting point is the agent's position. Moving along the global path at a sampling speed $v_{\text{samp}}$ for a time step $h$ results in arriving at the second reference point $\boldsymbol{p}^{l}_{1,\text{ref}}$. Sampling continues until $\boldsymbol{p}^{l}_{N,\text{ref}}$ is reached. The sampling speed $v_{\text{samp}}$ is adapted according to the density of the environment that the drone is navigating. The density can be inferred by whether the voxels that the global path traverses are in a potential field.

The sampling speed modulation algorithm allows for the agent to navigate faster in free environments and slower in cluttered environments. 
The algorithm takes as user input a minimum sampling speed $v_{\text{samp,min}}$ that the agent will have in extremely cluttered environments and a maximum sampling speed $v_{\text{samp,max}}$ that the agent will have in open/free space. The choice of both of those speeds is fine-tuned according to the dynamical limits of the agent.

\begin{algorithm}
\DontPrintSemicolon
\caption{Speed adaptation using the global path}
\label{alg:speed_adaptation}
 $v_{\text{samp}} = v_{\text{samp,max}}$\;
\For {$i$ = 1:size(path)}{
 $o_i = $ GetValue(path[$i$])\;
 $d_i = $ GetDistance(path[0], path[$i$])\;
 $\alpha = 1 - e^{-s_d\cdot d_i} \cdot (1-e^{-s_o\cdot o_i})$\;
 $v_{\text{tmp}} =  v_{\text{samp,min}} + \alpha \cdot (v_{\text{samp,max}} - v_{\text{samp,min}})$\;
  $v_{\text{samp}} = \min(v_{\text{samp}}, v_{\text{tmp}})$\;
}
\end{algorithm}

The algorithm for adapting the speed is described in Alg. \ref{alg:speed_adaptation}.
The sampling speed is first set to the maximum sampling speed $v_{\text{samp,max}}$ (line 1). We then walk along the global path starting from the first position (line 2). At every point/voxel of the path, we get the voxel value $o_i$ (line 3), which is set according to the equation (\ref{eqn:potential}) (we consider unknown voxels to be occupied). We also get the distance $d_i$ of the point to the starting point of the path, which is close to the current position of the agent (line 4). We then compute the scaling factor $\alpha$ that depends on $o_i$, $d_i$, and the user-chosen distance sensitivity $s_d$ and potential sensitivity $s_o$ (line 5). The sensitivities will determine how fast we slow down once we realize that the path is entering a potential field, and thus becoming close to obstacles. We compute the velocity limit set by the current point/voxel $v_{\text{tmp}}$ (line 6) and then set the sampling speed to $v_{\text{tmp}}$ if it is smaller than its current value (line 7). Note that if a voxel has a value $o_i = 0$ (i.e. is free), then $\alpha$ will be 1, and $v_{\text{tmp}}$ will be  $v_{\text{samp,max}}$. On the other hand, if $o_i > 0$, then the bigger that value, the closer we are to an obstacle (as per equation (\ref{eqn:potential})) i.e. the denser the environment. In that case, $\alpha$ will be smaller than 1 and the velocity will be smaller than $v_{\text{samp,max}}$. The distance of the point/voxel $d_i$ to the starting point of the path is also taken into account: the smaller that distance, the closer we are to obstacles and the smaller $\alpha$ is i.e. the smaller the sampling velocity is.

The reference path is sampled at each planning iteration. In subsequent iterations, and as the agent moves forward and progresses along the path, the reference trajectory needs to move along the global path as well. \textit{Path progress} is used as a measure of when to move along/increment the starting point $\boldsymbol{p}^{l}_{0,\text{ref}}$  of the sampling on the path.
    
\begin{figure}
\centering
\includegraphics[trim={0cm 0cm 0cm 0cm},clip,width=0.9\linewidth]{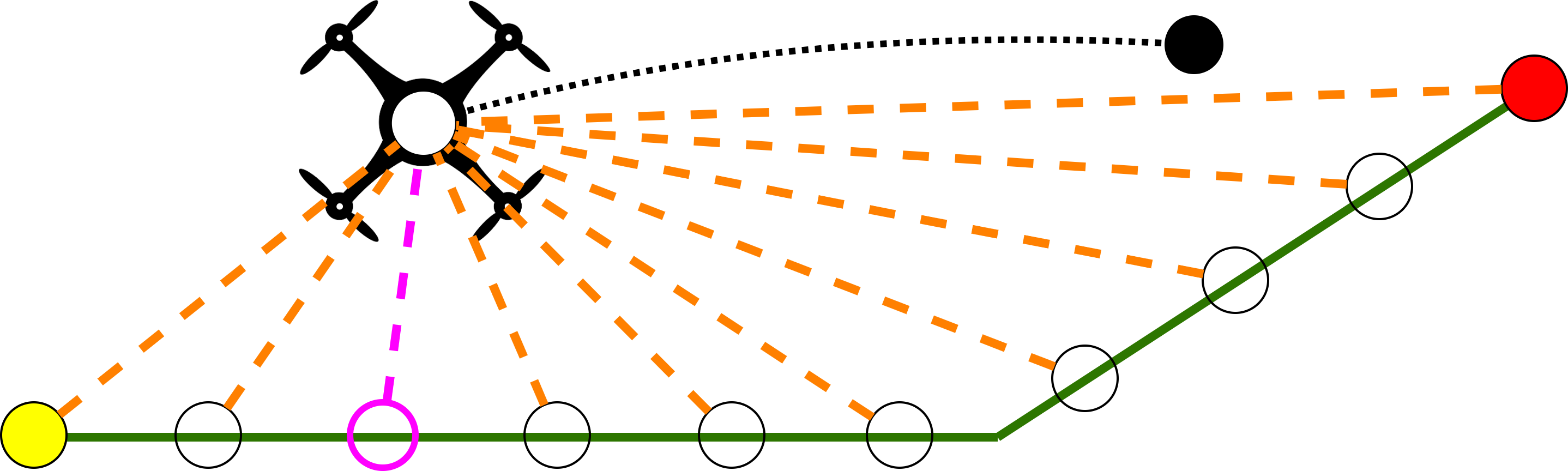}
\caption{The process of computing the path progress: the local reference trajectory (green path) is sampled (green circles) and we find the closest point among these samples to the drone (magenta circle). The path progress is the distance between the first point of the local reference trajectory (yellow circle) and the magenta circle. The predicted trajectory of the drone is the black dotted line and the final position is the black disk. The red disk is the final point in the local reference trajectory.}
\label{fig:path_progress}
\end{figure}

Path progress is computed in the following way (Fig. \ref{fig:path_progress}): first, the local reference trajectory is sampled by a small value (typically 1 cm). Then the point $\boldsymbol{p}_{\text{min}}$ (magenta circle in Fig. \ref{fig:path_progress}) that has the minimum distance to the agent among the sampled points is found. The distance between  $\boldsymbol{p}_{\text{min}}$ and the starting point (yellow disk in Fig. \ref{fig:path_progress}) of the local reference trajectory is the path progress. If at a given iteration $l$ the path progress is 0 i.e. the closest point is the starting point, then the same starting point for the sampling is kept  $\boldsymbol{p}^{l}_{0,\text{ref}} = \boldsymbol{p}^{l-1}_{0,\text{ref}}$. If the path progress is bigger than 0, then we move along the starting point of the sampling $\boldsymbol{p}^{l}_{0,\text{ref}} = \boldsymbol{p}^{l-1}_{1,\text{ref}}$.

The output of the path planning algorithm can vary in terms of direction between one planning iteration and another, which results in reference trajectories with a varying direction between iterations. This can result in jerky behavior as the drone is trying to follow a reference trajectory that is changing direction every iteration.
To fix this issue, the path planning module is given the last point of the reference trajectory as the initial position to plan from, which results in smoother navigation. The degree of smoothness can be optionally fine-tuned during the experiments by starting the path planning from a reference point between the current position and the last reference point (instead of starting from the last point). The index of this point is denoted $i_{\text{path,start}}$.

\subsubsection{MPC/MIQP solver}
\label{app:mpc_solver}
In this section, we will derive the MPC/MIQP formulation that is solved at every planning iteration to generate the collision-free optimal trajectory that is sent for the controller to execute.
The initial state $\boldsymbol{x}_{0}^l$ of the generated trajectory at iteration $l$ is set to the second state of the trajectory generated at the previous iteration $l-1$ i.e. $\boldsymbol{x}_{0}^l = \boldsymbol{x}_{1}^{l-1}$. The terminal velocity $\boldsymbol{v}_N$ as well as the terminal acceleration $\boldsymbol{a}_N$ are set to $\boldsymbol{0}$ to guarantee safety. If a trajectory optimization fails or the computation time exceeds the trajectory planning period $T_{\text{traj}}$, then the last successfully generated trajectory continues to be executed. Since the terminal velocity and acceleration of each agent is $\boldsymbol{0}$, if all trajectories fail indefinitely to be generated, the agents will come to a stop and will not collide.

The agent dynamics are represented by a point mass (center of gravity of the agent) that is controlled by the jerk $\boldsymbol{j}$. This representation can be used for omnidirectional vehicles in general, and multirotors in particular \cite{toumieh2022near}. We also introduce drag forces to ensure that the trajectory is still feasible at high speeds. The drag forces are linear with respect to the velocity to make sure that the dynamics are convex and the MPC solving time is low. The dynamics are the following:
\begin{gather}
  \label{eqn:cont_dynamics}
    \begin{aligned}
     & \Dot{\boldsymbol{p}} = \boldsymbol{v}, \quad
      \Dot{\boldsymbol{v}} = \boldsymbol{a} - \boldsymbol{D}_{\text{lin,max}}\cdot \boldsymbol{v}, \quad 
     \Dot{\boldsymbol{a}} = \boldsymbol{j}
    \end{aligned}
\end{gather}
where $\boldsymbol{p}$ is the position vector, $\boldsymbol{v}$ the velocity vector, $\boldsymbol{a}$ the acceleration vector, and $\boldsymbol{D}_{\text{lin,max}}$ a diagonal matrix that represents the maximum possible drag coefficient in every direction.

The continuous dynamics are discretized using Euler or Runge-Kutta 4\textsuperscript{th} and added as equality constraints to the optimization problem.
The state of each agent is formed by the position, velocity, and acceleration $\boldsymbol{x} = [\boldsymbol{p} \ \boldsymbol{v} \ \boldsymbol{a}]^T $ and the control input corresponds to the jerk $\boldsymbol{u} = \boldsymbol{j}$. The continuous dynamics can be written under the form
 $\dot{\boldsymbol{x}} =   f(\boldsymbol{x}(t),\boldsymbol{u}(t))$. We choose Euler discretization due to its better time efficiency and discretize the dynamics that become under the form $\boldsymbol{x}_{k+1} = f_d(\boldsymbol{x}_k,\boldsymbol{u}_k)$. Using a discretization step of $h$, the discretized dynamics are:
\begin{equation}
    \begin{aligned}
     & \boldsymbol{p}_{k+1} = \boldsymbol{p}_{k} + h\cdot \boldsymbol{v_k} \\
     & \boldsymbol{v}_{k+1} = \boldsymbol{v}_{k} + h\cdot ( \boldsymbol{a}_{k} -  \boldsymbol{D}_{\text{lin,max}}\boldsymbol{v_k}) \\
     & \boldsymbol{a}_{k+1}= {a}_{k} + h\cdot \boldsymbol{j_k} \\
     & \boldsymbol{x}_{k} = [\boldsymbol{p}_{k} \  \boldsymbol{v}_{k} \ \boldsymbol{a}_{k}]^T, \quad \boldsymbol{u}_k = \boldsymbol{j}_k
    \end{aligned}
\end{equation} 

State bounds are inequality constraints that are added to the optimization to ensure that the generated trajectory is dynamically feasible. The agent's physical dynamics and limitations determine the bounds that are set to its acceleration and jerk. The velocity of each agent is inherently limited by the drag forces in the dynamics. We limit the L1 norm of the acceleration in the directions $x$ and $y$ to $a_{x,\text{max}}$ and $a_{y,\text{max}}$ respectively. We limit the acceleration in the $z$ direction to $a_{z,\text{min}}$ and $a_{z,\text{max}}$ since it is harder to accelerate upwards (fighting gravity) than downwards. The acceleration limitations are deduced from the thrust-to-weight ratio of the multirotor. The L1 norm of the control input (jerk) is limited in each direction to $j_{x,\text{max}}$, $j_{y,\text{max}}$ and $j_{z,\text{max}}$. These limitations are determined by the rotational dynamics of the multirotor.

The generated trajectory should avoid static obstacles as well as other agents. For this end, it is contained inside the TASC generated in Sect. \ref{sect:tasc}. The constraints are generated by forcing 2 consecutive positions $\boldsymbol{p}_k$ and $\boldsymbol{p}_{k+1}$ (and thus the segment formed by them) to be inside one polyhedron of the time-local safe corridor $SC_k$. Each $SC_k$ incorporates constraints based on the positions of all agents at discrete time points $k$ in the future. This enables the planning agent to plan within the anticipated free space at time $h\cdot k$. Each $SC_k$ contains $P_{\text{hor}}$ polyhedra each described by
$\{(\boldsymbol{A}_{kp}, \boldsymbol{c}_{kp})\}$, $p = 0 : P_{\text{hor}}-1$.The constraint ensuring that the discrete position, represented by $\boldsymbol{p}_k$, resides within a polyhedron $p$ is expressed as $\boldsymbol{A}_{kp} \cdot \boldsymbol{p}_k \leq \boldsymbol{c}_{kp}$. Binary variables $b_{kp}$ are introduced (with $P_\text{hor}$ variables for each $\boldsymbol{x}_k$, where $k = 0 : N-1$) to indicate whether $\boldsymbol{p}_k$ and $\boldsymbol{p}_{k+1}$ are both within polyhedron $p$. The requirement that all segments be within at least one polyhedron is enforced by the constraint $\sum_{p=0}^{P-1} b_{kp} \geq 1$. This necessitates the existence of an overlapping region across all polyhedra, with at least one discrete point belonging to this overlap when transitioning from one polyhedron to another.
In our framework, we limit the number of polyhedra per Safe Corridor $P_{\text{hor}}$ to 4 to avoid high solving times.

The cost function consists of the squared norm of the difference between the predicted state $\boldsymbol{x}_k$ and its reference $\boldsymbol{x}_{k,\text{ref}}$ ($k = 1, 2, ..., N$). Each reference $\boldsymbol{x}_{k,\text{ref}}$ is created from the reference trajectory position:
\begin{gather}
    \boldsymbol{x}_{k,\text{ref}} = [\boldsymbol{p}_{k,\text{ref}} \ \boldsymbol{v}_{k,\text{ref}} \ \boldsymbol{0}_{1\times3}]^T
\end{gather}
where $\boldsymbol{v}_{k,\text{ref}}$ is a vector of norm $v_{\text{samp}}$ and of direction $\boldsymbol{v}_{\text{dir}} =  \boldsymbol{p}_{k+1,\text{ref}} - \boldsymbol{p}_{k,\text{ref}}$ (except for $\boldsymbol{v}_{N,\text{ref}}$ which is equal to $\boldsymbol{v}_{N-1,\text{ref}}$). The control input (jerk) is added to the cost function to guarantee a level of smoothness.

We express our Model Predictive Control (MPC) using the Mixed-Integer Quadratic Program (MIQP) formulation. To simplify, we omit the superscript $l$ denoting the iteration number from both the reference and state variables:
\begin{align}
    & \underset{\substack{\boldsymbol{x}_k,\boldsymbol{u}_k}}{\text{minimize}}
& & \sum_{k=0}^{N} (||\boldsymbol{x}_k - \boldsymbol{x}_{k,\text{ref}}||_{\boldsymbol{R}_x}^2 +  ||\boldsymbol{u}_k||_{\boldsymbol{R}_u}^2) \nonumber \\
& & & + ||\boldsymbol{x}_N - \boldsymbol{x}_{N,\text{ref}}||_{\boldsymbol{R}_N}^2 \label{eq:NLP}\\
& \text{subject to} & & \boldsymbol{x}_{k+1} = f_d(\boldsymbol{x}_k,\boldsymbol{u}_k),\quad k = 0:N-1 \label{eq:k}\\
& & & \boldsymbol{x}_0 = \boldsymbol{X}_0, \quad \boldsymbol{v}_N = \boldsymbol{0}, \quad \boldsymbol{a}_N = \boldsymbol{0}  \\
& & & |a_{x,k}| \leq a_{x,\text{max}} \\
& & & |a_{y,k}| \leq a_{y,\text{max}}, \quad a_{z,k} \leq a_{z,\text{max}} \\
& & & a_{z,k} \geq a_{z,\text{min}}, \quad |j_{x,k}| \leq j_{x,\text{max}} \\
& & & |j_{y,k}| \leq j_{y,\text{min}}, \quad |j_{z,k}| \leq j_{z,\text{max}} \\
& & & b_{kp} = 1 \implies \begin{cases} \boldsymbol{A}_{kp}\cdot \boldsymbol{p}_k \leq \boldsymbol{c}_{kp} \\
\boldsymbol{A}_{kp}\cdot \boldsymbol{p}_{k+1} \leq \boldsymbol{c}_{kp} \label{eqn:const_poly}
\end{cases}\\
& & & \sum_{p=0}^{P_{\text{hor}}-1} b_{kp} \geq 1, \quad b_{kp} \in \{0,1\}
\end{align}

$\boldsymbol{R}_x$, $\boldsymbol{R}_N$, and $\boldsymbol{R}_u$ represent the weight matrices for discrete state errors excluding the final state, the final discrete state error (terminal state), and the input, respectively.

The optimization problem is solved in each planning iteration using the Gurobi solver \cite{gurobi} to produce an optimal trajectory based on its cost function.

\subsection{Packet loss}
\label{app:packet_loss}
If an agent does not receive the trajectory of at least one nearby agent before its planning for the next iteration begins, it will keep executing its last generated trajectory. This guarantees that no collisions will occur because of the following reasoning. Let's consider the case of 2 agents A and B. If agent A does not receive the trajectory of agent B at a given planning iteration, it will continue executing its previous trajectory of fixed time horizon. Since all generated trajectories have a terminal state with zero velocity and acceleration, the agent will be stationary when it reaches the end of the trajectory. We have 2 cases: either agent B also did not receive the trajectory of agent A, or it did receive it. \begin{itemize}
    \item If agent B did not receive the trajectory of agent A, it also continues executing its previously generated trajectory, and safety is guaranteed because both trajectories were optimized using the same separating hyperplanes at the previous planning iteration. In the extreme case where both agents lose complete communication, they will execute their last generated trajectory until they reach the terminal state and remain stationary until communication is re-established.
    \item If agent B received the trajectory of agent A, new hyperplanes are generated using the trajectory that agent A is currently executing and the last generated trajectory of B. The key point is that we will always be able to generate hyperplanes that guarantee collision safety because all previous trajectories from which the hyperplanes are generated have safety guarantees i.e. the positions of the agents are always more distant than the safety distance due to the previous hyperplanes.
\end{itemize} 

\bibliographystyle{IEEEtran}
\bibliography{IEEEabrv,IEEEexample}

\vspace{11pt}

\begin{IEEEbiography}[{\includegraphics[width=1in,height=1.25in,clip,keepaspectratio]{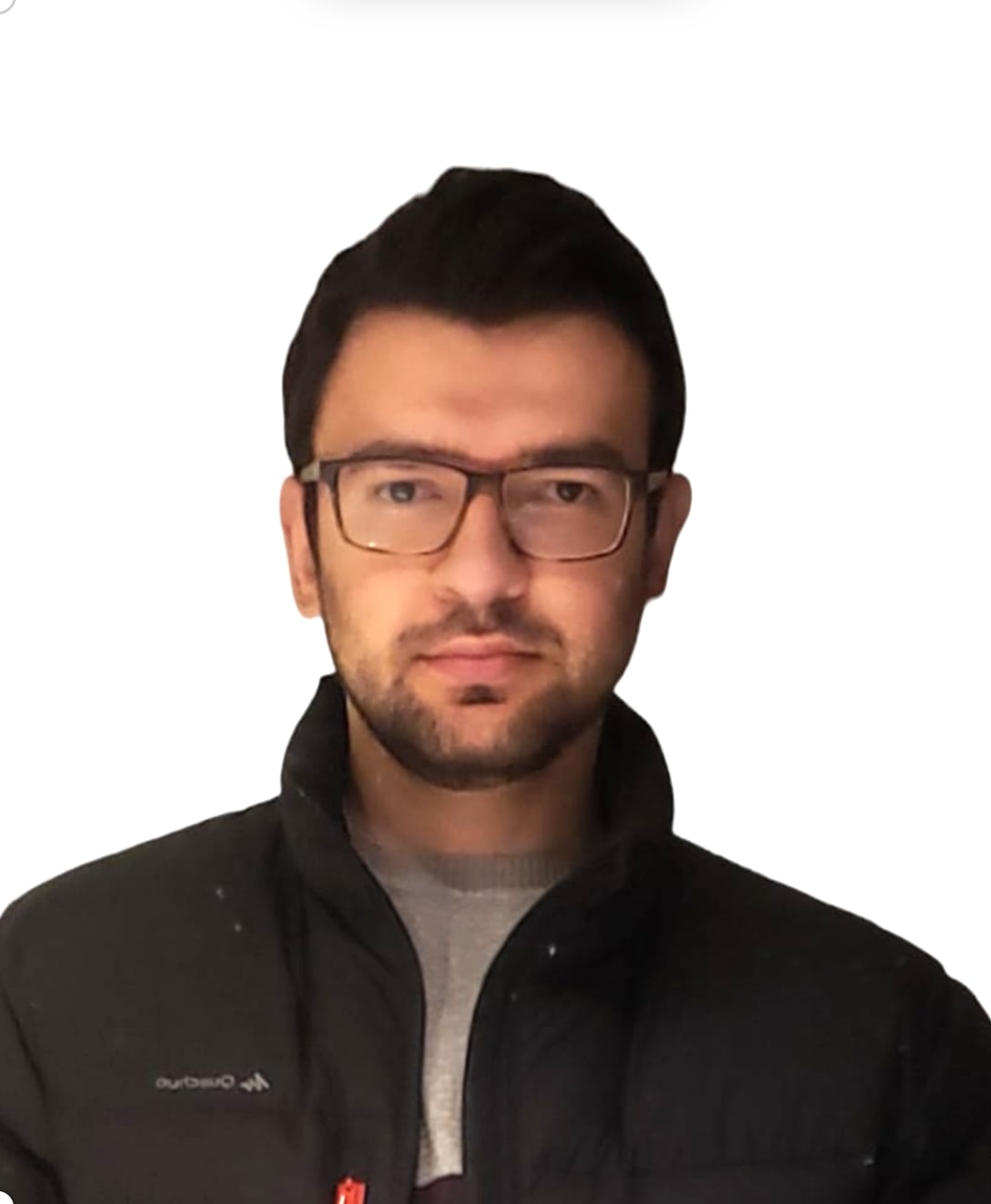}}]{Charbel Toumieh} is a postdoctoral fellow at the Ecole Polytechnique Federale de Lausanne, Lausanne, Switzerland. He received his Ph.D. in robotics from Paris-Saclay University, France in 2022. He received his M.Sc. in Electrical and Computer 
Engineering from the Georgia Institute of Technology, USA in 2019. His 
research interests include autonomous robotics, motion planning, 
multi-agent systems, and aerial swarms.
\end{IEEEbiography}
\begin{IEEEbiography}[{\includegraphics[width=1in,height=1.25in,clip,keepaspectratio]{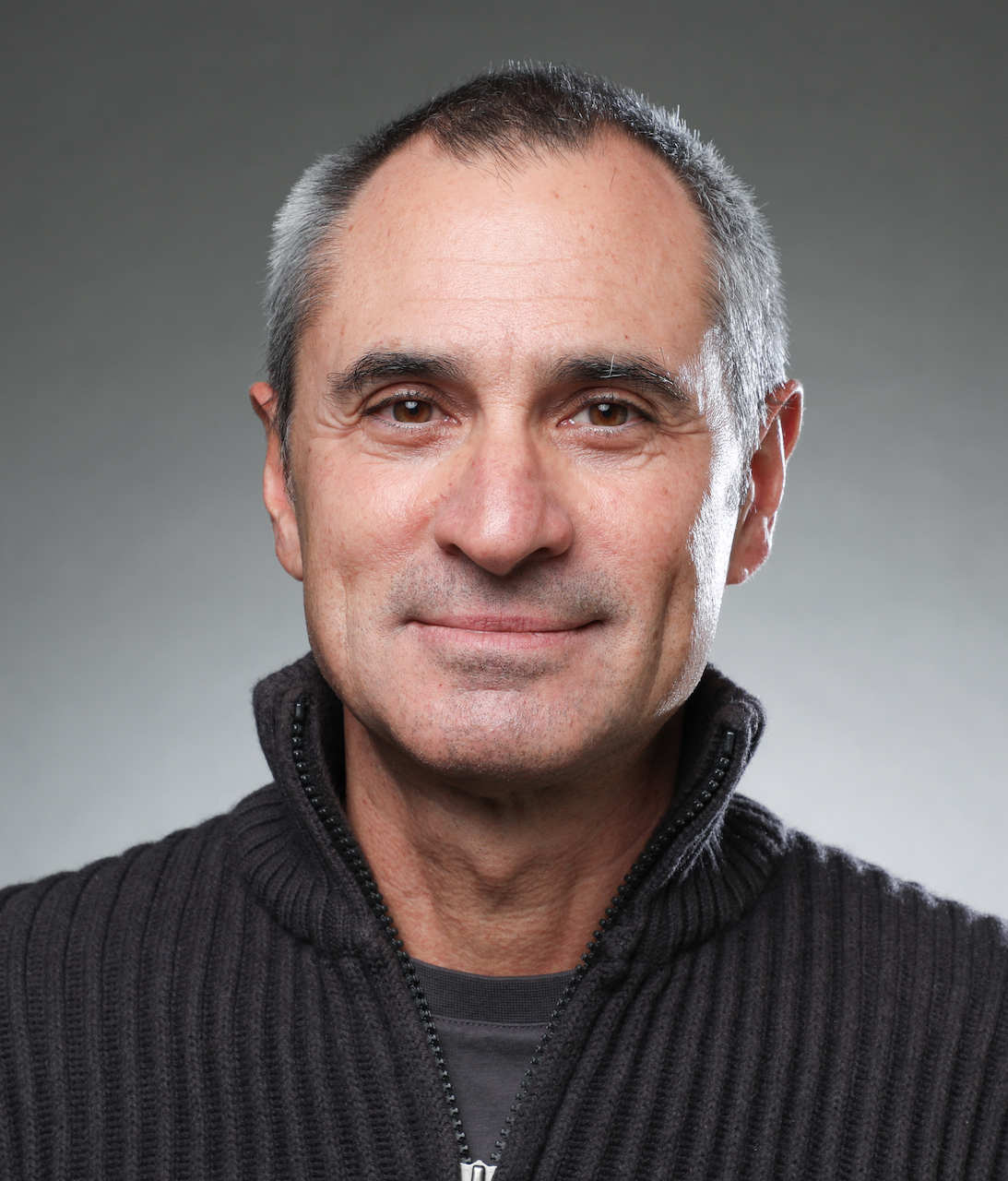}}]{Dario Floreano} (Fellow, IEEE) received
the M.A. degree in vision psychophysics, the M.S. degree in neural computation, and the Ph.D. degree in robotics. 

He is currently a Full Professor and Director of
the Laboratory of Intelligent Systems, Swiss Federal Institute of Technology Lausanne (EPFL), Lausanne,
Switzerland. Since 2010, he is also a Founding Director of the Swiss National Center of Competence in Robotics. He held Visiting Fellowships with Sony Computer Science Laboratory, Tokyo, Japan, with
Caltech/JPL, Pasadena, CA, USA, and with Harvard University in Boston, Boston, MA, USA. His research interests focus on biologically inspired robotics
and artificial intelligence. He made pioneering contributions to the fields of
evolutionary robotics, aerial robotics, and soft robotics. He spun off two robotics
companies: senseFly (2009, acquired by the Parrot Group in 2016), which has
become a world leader in drones for agriculture and imaging, and Flyability
(2015), which is the world leader in inspection drones for confined spaces. In
2017, The Economist dedicated a center page portrait to Prof. Floreano (Brain
Scan).

Prof. Floreano served in the Advisory Board of the Future and Emerging
Technologies division of the European Commission, has been Vice-Chair of
the World Economic Forum Agenda Council on Smart Systems and Robotics,
has been a Cofounder of the International Society of Artificial Life, served as
elected member of the Board of Governors of the International Neural Network
Society, is on the Advisory Board of the Max-Planck Institute for Intelligent
Systems, and has joined the editorial board of ten scientific journals. He also
co-organized ten international conferences and several thematic workshops on
bioinspired drones and soft robotics, which are now considered foundational
events for those communities
\end{IEEEbiography}
\vfill

\end{document}